\documentclass[conference]{IEEEtran}
\usepackage[ruled,linesnumbered,noend,noline]{algorithm2e}
\usepackage{url}
\usepackage{amsmath,amssymb,amsfonts}
\usepackage{graphicx}
\graphicspath{{figures/}}
\newcommand{\RNum}[1]{\uppercase\expandafter{\romannumeral #1\relax}}
\newif\ifsubmission
\submissiontrue

\ifsubmission
\newcommand{\note}[1]{}
\newcommand{\eat}[1]{}
\newcommand{\ygedit}[1]{{}}
\newcommand{\ygcomment}[1]{}
\newcommand{\yycomment}[1]{}
\newcommand{\cscomment}[1]{}
\newcommand{\hzcomment}[1]{}
\newcommand{\todo}[1]{}
\newcommand{\ygtodo}[1]{}
\newcommand{\hztodo}[1]{}

\newcommand{\done}[1]{}
\else
\newcommand{\note}[1]{#1}
\newcommand{\eat}[1]{}
\newcommand{\ygedit}[1]{{\color{blue} #1}}
\newcommand{\ygcomment}[1]{\ygedit{[YG: #1]}}
\newcommand{\yycomment}[1]{\ygedit{[YQ: #1]}}
\newcommand{\cscomment}[1]{\ygedit{[CS: #1]}}
\newcommand{\hzcomment}[1]{\ygedit{[HY: #1]}}
\newcommand{\todo}[1]{\textcolor{red}{\emph{TODO: #1}}}
\newcommand{\ygtodo}[1]{\textcolor{red}{\emph{TODO (yaguang): #1}}}
\newcommand{\hztodo}[1]{\textcolor{red}{\emph{TODO (haoze): #1}}}

\newcommand{\done}[1]{\textcolor{red}{\st{\emph{TODO: #1}} \textcolor{green}{[Done]}}}

\fi

\newtheorem{definition}{Definition}

\usepackage{soul}
\usepackage{algorithmic}
\usepackage{booktabs}
\usepackage{subcaption}
\usepackage{enumitem}
\usepackage{siunitx}
\usepackage{cite}

\makeatletter
\def\ps@IEEEtitlepagestyle{%
  \def\@oddfoot{\mycopyrightnotice}%
  \def\@evenfoot{}%
}
\def\mycopyrightnotice{%
  {\footnotesize 978-1-7281-0858-2/19/\$31.00 © 2019 IEEE \hfill}
  \gdef\mycopyrightnotice{}
}

\begin{document}

\title{DETECT: Deep Trajectory Clustering for Mobility-Behavior Analysis}
\author{\IEEEauthorblockN{Mingxuan Yue, Yaguang Li, Haoze Yang, Ritesh Ahuja, Yao-Yi Chiang, Cyrus Shahabi}
\IEEEauthorblockA{University of Southern California\\
       Los Angeles, U.S.A\\
       \{mingxuay, yaguang, haozeyan, riteshah, yaoyic,  shahabi\}@usc.edu}}

\maketitle

\begin{abstract} 
Identifying mobility behaviors in rich trajectory data is of great economic and social interest to various applications including urban planning, marketing and intelligence.
Existing work on trajectory clustering often relies on similarity measurements that utilize raw spatial and/or temporal information of trajectories. These measures are incapable of identifying similar moving behaviors that exhibit varying spatio-temporal scales of movement. In addition, the expense of labeling massive trajectory data is a barrier to supervised learning models. To address these challenges, we propose an unsupervised neural approach for mobility behavior clustering, called the Deep Embedded TrajEctory ClusTering network (DETECT). DETECT operates in three parts: first it transforms the trajectories by summarizing their critical parts and augmenting them with context derived from their geographical locality (e.g., using POIs from gazetteers). In the second part, it learns a powerful representation of trajectories in the latent space of behaviors, thus enabling a clustering function (such as $k$-means) to be applied. Finally, a clustering oriented loss is directly built on the embedded features to jointly perform feature refinement and cluster assignment, thus improving separability between mobility behaviors. Exhaustive quantitative and qualitative experiments on two real-world datasets demonstrate the effectiveness of our approach for mobility behavior analyses.
\end{abstract}

\maketitle

\section{Introduction} \label{sec:intro}
The rapid proliferation of  mobile devices has led to the collection of vast amounts of GPS trajectories by location-based services, geo-social networks and ride sharing apps. 
In these trajectories, sophisticated human \textit{mobility behaviors} are encoded. We define \textit{mobility behavior} (of a trajectory) as the travel activity that describes a user's movements regardless of the spatial and temporal distances that he covers. For example, work-to-home commute is one such mobility behavior that varies widely in terms of the area-travelled and the time-taken to complete the activity.

The analysis of mobility behavior is of tremendous value to various applications. For instance, recommending a point-of-interest (POI) for a user to visit next is an important task in mobile applications. Given that latent factors describing mobility behaviors can facilitate location-based advertisement \cite{learning2017zhu}, a number of existing approaches successfully exploit this knowledge for next-POI recommendation. Most~\cite{he2016inferring,chang2018content} involve extracting some form of latent patterns within user trajectories. Besides recommender-systems, detecting mobility behaviors in trajectory data enable policymakers and law enforcement to better allocate resources and detect abnormal or dangerous activities~\cite{zheng2015trajectory, chaoyang}.

While latent factors that supposedly represent mobility behaviors have been used to good effect, to the best of our knowledge, there is no existing work that facilitates the detection and analysis of mobility behaviors in trajectory data. 
At its core, the bottleneck to studying mobility behaviors is the absence of labeled data that can support supervised learning, especially for the large data volumes consisting of users' trajectories. Manually labeling each trajectory with its mobility behavior is prohibitively expensive and requires expert skills, making unsupervised learning methods highly desirable. Hence we formulate the problem of mobility-behavior analysis as a \emph{clustering} task. There are mainly two challenges to obtaining a good clustering:\\

\noindent\textbf{Challenge 1: Scale-variance.} Trajectories with similar mobility behavior exhibit various spatial and/or temporal range of movement. For example, a ``commute to work'' behavior could take from 10 minutes to up to 1 hour via different transportation modes. Even with the same transportation modality, the travel-time fluctuates at different times of the day (e.g., rush hours), needing us to account for varying temporal scales within the same mobility behavior. Likewise, the spatial range of movement also varies, e.g., from a mile to more than forty, for  users living near or far away from their workplaces.

Traditional trajectory clustering techniques (e.g.,~\cite{lee2007trajectory, yoon2008robust, yuan2017review, lin2008one}) group trajectories based on raw spatial and temporal distances that are sensitive to variation in the spatio-temporal scale. These methods fail to cluster mobility behaviors, and instead produce simple clusters, each with similar spatio-temporal range of movement. This is a direct consequence of the Euclidean distance metric being incompatible with length-variant and misaligned trajectories~\cite{liao2005clustering}. Recent methods borrow from work in clustering time-series data; utilizing distance measurements better suited to flexible alignments between trajectories, such as Dynamic Time Warping (DTW), Longest Common SubSequence (LCSS) and the Fr\'echet distance. While alignment-based distance measurements produce modest improvements over the Euclidean distance metric, the solutions are not satisfying since they allow limited temporal disordering, possess high sensitivity to outliers, and heavily rely on local similarities~\cite{yuan2017review}. \\\\

\noindent\textbf{Challenge 2: Lack of context.} Mobility behavior of a trajectory is highly dependent on characteristics of places and environments in which it occurs, henceforth called the \textit{geographical context}. For example, the ``grocery shopping'' behavior encompasses trajectories with very different travel times and shapes depending on the user's choice of a supermarket. However, in this instance, knowing the geographical context, i.e. the supermarket, is more informative than the exact shape and duration of the trajectory. It is therefore necessary to integrate geographical influence at each point of the trajectory to achieve good clustering quality.
In brief, a good clustering framework must learn transitions of the geographical context along the trajectory, while remaining insensitive to the spatial and temporal range of movement.\\

\noindent\textbf{Approach.} To reconcile these challenges, we propose the Deep Embedded TrajEctory ClusTering network (DETECT), as a unified framework to cluster trajectories according to their mobility behaviors. DETECT operates in three parts: first, it summarizes the critical parts of the trajectory and augments them with context derived from their geographical locality (e.g., using POIs from gazetteers). The augmented trajectories now incorporate semantics essential for identifying mobility behaviors, but are still of variable lengths. In the second part, DETECT handles variable-length input by adapting an  autoencoder~\cite{kingma2013auto}, trained over a large volume of unlabeled trajectories. The autoencoder instantiates an architecture for learning the distributions of latent random variables to model the variability observed in trajectories. On the learned embedding in the low-dimensional feature space, a clustering function (e.g., $k$-means) is applied. In the last part---as the most computationally intensive procedure of DETECT---a clustering oriented loss is directly built on embedded features to jointly perform embedding refinement and cluster assignment. The joint optimization iteratively and finely updates the embedding towards a high confidence clustering distribution by improving separability between mobility behaviors. In summary, this paper has the following major contributions:
\begin{itemize}
\item We propose a powerful unsupervised neural framework to cluster GPS trajectories for the problem of mobility behavior detection, while addressing scale-variance and context-absence of raw GPS points.
\item We propose a novel feature augmentation process for mobility behavior analysis that augments the GPS trajectories with effective input features; that, by providing mobility context, can even improve the performance of baseline approaches.
\item We propose a neural network architecture that takes as input the augmented trajectory features to embed in a fixed-length latent space of behaviors, and gradually improves the embedding for a better clustering.

\item We conduct an extensive quantitative and qualitative evaluation on real-world trajectory datasets to show that the proposed approach outperforms the state-of-the-art approaches significantly; ranging from at least 41\%, and up to 252\% across well-established clustering metrics.
\end{itemize}
The remainder of the paper is organized as follows. We describe DETECT in Section~\ref{sec:detect}. Section~\ref{sec:exp} gives an exhaustive experimental evaluation. Finally, Section~\ref{sec:relat} presents the related work and Section~\ref{sec:con} concludes.


\section{DETECT} \label{sec:detect} 

\noindent\textbf{Problem.} We define mobility behavior clustering as the problem of grouping trajectories in such a way that trajectories in the same group have similar transitions of travel activity context. \\

\noindent\textbf{Data.} The input is a set of raw trajectories, where each trajectory $s = \{s^{(1)}, s^{(2)},$ $ \ldots, s^{(T)}\}$ is a time-ordered sequence of spatio-temporal points. Each point $s^{(t)}$ consists of a pair of spatial coordinates (i.e. latitude, longitude) and its timestamp. A spatio-temporal point is often surrounded by Points-Of-Interest (POI), which represent locations someone may find useful or interesting, such as a high school, a business office or a wholesale store. We represent a POI with a spatial coordinate and a major category of service $m \in M$ (e.g, education, commerce or shopping). \\

\noindent\textbf{Overview.} DETECT is comprised of three parts. In the first part (Section~\ref{subsec:feature}), DETECT operates over the input raw trajectories, and summarizes the critical parts of the trajectory as \textit{stay points}, incurring little loss of information. Stay points form ideal candidates to discover the context of the travel activity through a representation of their geographical locality as feature vectors. In the next part, labeled DETECT Phase~\RNum{1} (Section~\ref{subsec:phase1}), the context augmented trajectories are embedded into a latent space of mobility behaviors. The low-dimensional latent space enables a simple clustering function to be then applied. Finally in the third part (Section~\ref{subsec:phase2}), labeled DETECT Phase~\RNum{2}, the clustering assignment is refined by updating the latent embedding via a clustering oriented loss. 



\subsection{Feature Augmentation}\label{subsec:feature}
\subsubsection{Stay point detection} 
\vspace{-3pt}
Trajectories widely vary in their spatial and temporal range of movements. For each trajectory, the numerous GPS points recorded generally do not contribute significant information to support the detection of mobility behavior; which tends to be correlated with the context of the stops where the actual activity occurs. We argue the same empirically in Section~\ref{subsec:ablation}. In this work, we explore stay point detection~\cite{li2008mining} as a type of \textit{trajectory summarization} customized to reducing the spatio-temporal scale in trajectories. First introduced in~\cite{li2008mining} for discovering location-embedded social structure, stay points also offer several benefits to neural learning on trajectories. Due to their condensed format, they improve learning efficiency and  enable the neural layer to learn a better representation. 

\begin{definition}[Stay Points~\cite{li2008mining}]
A stay point $\dot s^{(t)}$ of trajectory $s$ is a spatiotemporal point, which is the geometric center of a longest sub-trajectory $s^{(i\rightarrow j)}\subset{s}, s^{(i\rightarrow j)} = \{s^{(i)}, s^{(i+1)}, \ldots, s^{(j)}\}$, such that $s^{(i\rightarrow j)}$ is a staying subtrajectory, and neither $s^{(i-1\rightarrow j)}$ nor $s^{(i\rightarrow j+1)}$ is a staying subtrajectory.
\end{definition}

In the original work, the algorithm to extract stay points (denoted SPD) has a computational complexity of $O(T^2)$ ($T$ is the length of the trajectory), which does not scale to real-world lengthy trajectories. We propose Fast-SPD, an efficient algorithm for stay point detection that exploits spatial indexes (e.g., R-Tree) to reduce the search space and quickly identify trajectory points close in space and time. Fast-SPD relies on efficient parsing of \textit{Staying Subtrajectories} defined as
\begin{definition}[Staying Subtrajectory]
A staying subtrajectory $s^{(i\rightarrow j)}\subset{s}, s^{(i\rightarrow j)} = \{s^{(i)}, s^{(i+1)}, \ldots, s^{(j)}\}$ of trajectory $s$ is a contiguous sub-sequence of $s$, such that within $s^{(i\rightarrow j)}$, the trajectory is limited to a range $\rho_{s}$ in space, and its duration $s^{(j)}.time-s^{(i)}.time$ is longer than a specific threshold $\rho_{t}$.
\end{definition}

\begin{algorithm}[ht]
	\SetAlgoLined
	\DontPrintSemicolon
	\LinesNumbered
	    $i \gets 0$ , $\dot s \gets \{\} $\;
	    \While{$i < \text{\upshape len}(s) - 1$} {
	      \If{$\text{\upshape dist}(s^{(i)}, s^{(i+1)}) > \rho_{s}$} {
	        $i \gets i + 1$ ; \textbf{continue}\;
	      }
	      $\text{cands} \gets s^{(i+1 \rightarrow T)} \cap b( s^{(i)}, \rho_{s})$ \;
	      $\text{neighbors} \gets \text{CommonIdx}(\text{cands}, s^{(i+1 \rightarrow T)})$\;
	      \If{$\text{ \upshape last}(\text{\upshape neighbors}).time - s^{(i)}.time > \rho_{t}$} {
	        $\text{neighbors} \gets \text{neighbors} \cup \{s^{(i)}\}$\;
	        $\dot s \gets s \cup \{ \text{average}(\text{neighbors})\}$\;
	        $i \gets i + \text{len}(\text{neighbors})$\;
	      }
	       $i \gets i + 1$\;
	    }
	    $\dot s \gets \{s^{(0)}\} \cup \dot s \cup \{s^{(T)}\}$ \;
		\Return{$\dot s$}
\caption{Fast-SPD($s,\rho_s, \rho_t$)~\label{alg:fast_spd}}
\end{algorithm}

\begin{figure}
\centering
\begin{subfigure}[t]{.52\columnwidth}
  \centering
  \raisebox{7pt}{\includegraphics[width=\columnwidth]{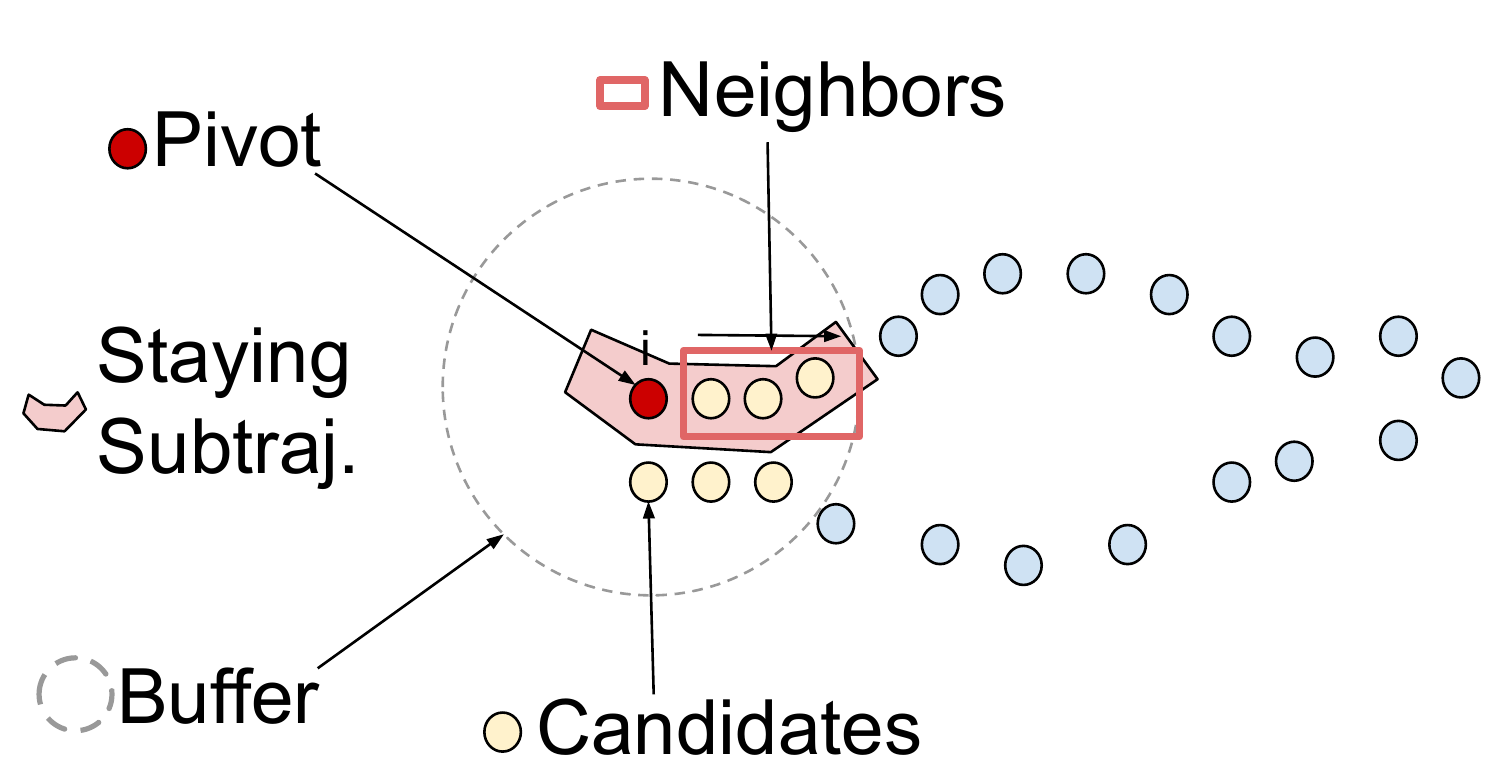}}
  \caption{SPD illustration}
  \label{fig:spd}
\end{subfigure}\hfill%
\begin{subfigure}[t]{.44\columnwidth}
  \centering
  \includegraphics[width=\columnwidth]{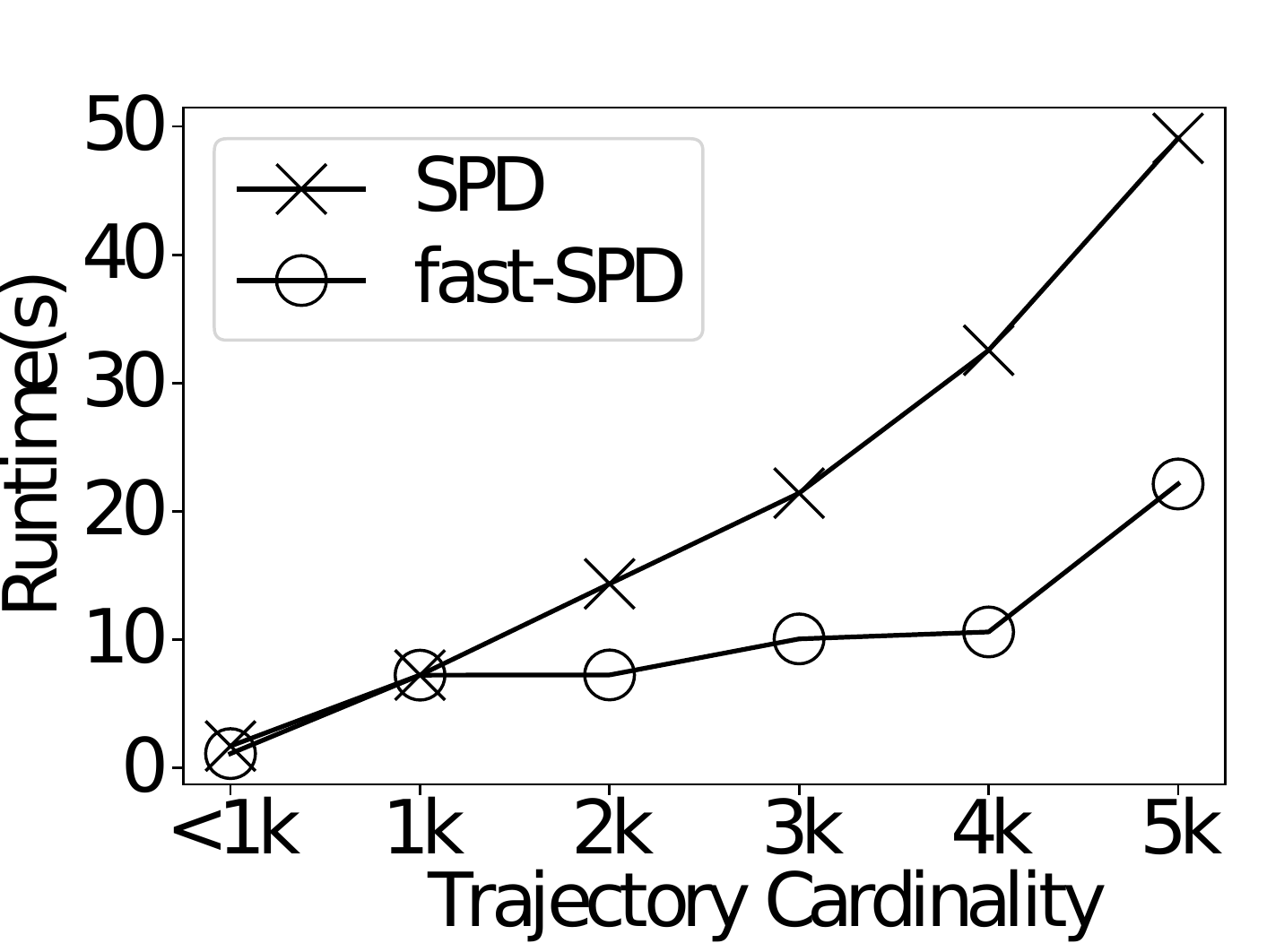}
  \caption{Running time}
  \label{fig:spdtime}
\end{subfigure}
\caption{Accelerated Stay Point Detection: Fast-SPD}
\vspace{-7pt}
\label{fig:test}
\end{figure}

The pseudocode of Fast-SPD is presented in Algorithm~\ref{alg:fast_spd}. In brief the algorithm, iterates through a trajectory point-by-point, finds all staying subtrajectories, and then generates stay points from each. We elaborate the procedure for stay point detection with an example in Figure~\ref{fig:spd}. Beginning with a pivot point $s^{(i)}$ (depicted as a red point) to search a staying subtrajectory. It first filters the candidate points (yellow points) by joining a spatial buffer (dotted circle) with the remaining points of the trajectory. Among the identified candidates, the consecutive points are refined to be termed  as neighbors (yellow points in the red-border box). If these neighbors cover a long enough period in time, i.e., $last(neighbors).time - s^{(i)}.time > \rho_{t}$, then these points along with the pivot point $s^{(i)}$ together constitute a staying subtrajectory (the pink box). Lastly, the algorithm extracts the geometric center of the staying subtrajectory as a stay point. Fast-SPD then skips the visited points and continues scanning the rest of the trajectory.

Compared to the SPD, Fast-SPD reduces the time complexity by $O(c)-O(log(T))$ for each search of a stay point, where $c$ (usually $c>log(T)$) is the length of a staying subtrajectory, and $T$ is the length of the trajectory.  Figure~\ref{fig:spdtime} depicts the running time of stay point detection with respect to the cardinality of the trajectory. Fast-SPD does not suffer on short trajectories ($\sim$1k points) but is up to three times more efficient on trajectories with many GPS points.

\noindent\subsubsection{Augmenting Geographical Context} \label{subsub:geocontext}
Given the extracted stay points, the next step is to incorporate the context of the surrounding environment (called geographical context) at each stay point.  The Tobler's First Law of Geography reported that "Everything is related to everything else, but near things are more related than distant things" \cite{tobler1970computer}. This phenomenon forms the basis for the study of Geographical Influences, which has seen wide adoption in the literature: from POI recommendation~\cite{zhang2014igeorec,yuan2013time,yu2015survey} to air quality prediction~\cite{lin2017mining}. In our work, the geographical context is constituted of categories (e.g., restaurants, apartments, hospitals) of the points of interest in the local vicinity of a stay point. DETECT transforms each point into a \textit{geographical context feature vector} defined as
\begin{definition}[Geographical Context Features]
The geographical context features of a stay point $\dot s^{(t)}_i$ are represented as a vector $x^{(t)}_i=\{x^{(t)}_{i,1}, x^{(t)}_{i,2}, \ldots, x^{(t)}_{i,M}\}$, where a feature $x^{(t)}_{i,m}$ denotes the contribution of the $m_{th}$ POI major category. 
\end{definition}
 
\begin{figure}[hb!]
\centerline{\includegraphics[width=0.7\columnwidth]{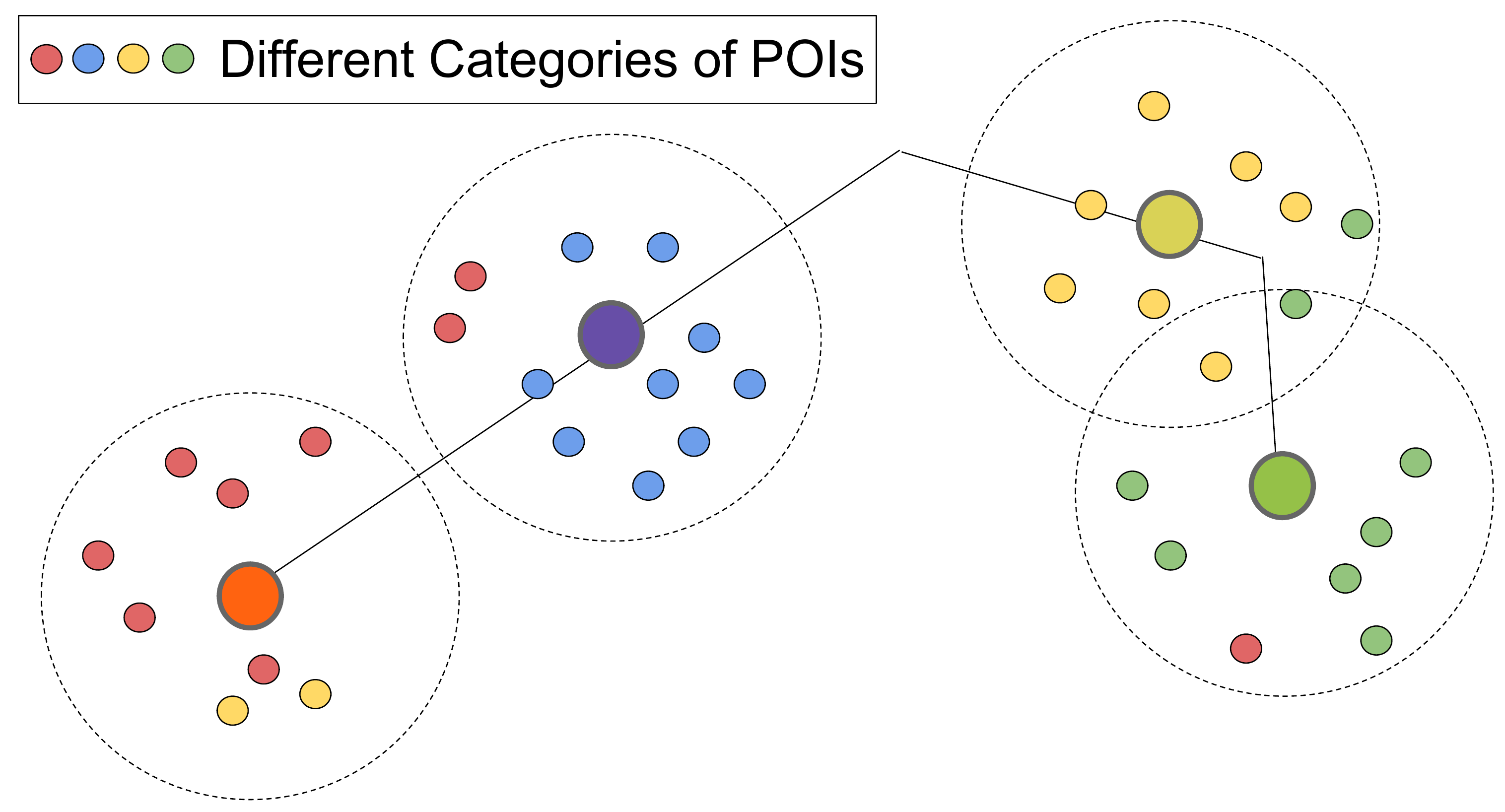}}
\caption{Augment stay points with geographical features}
\label{fig:poi}
\end{figure}

We briefly illustrate (in Figure~\ref{fig:poi}) the procedure to develop the feature vector from information existent in the locality of a stay point. For each point $\dot s^{(t)}$ (large solid points) in a stay point trajectory $\dot s = \{\dot s^{(t)}\}_{t=1}^T$, we execute a range search (circles) of radius $r_{poi}$ centered at the stay point. Next, within each circle, we summarize the counts of POIs in every category (small points in different colors) and normalize them by the total count within the circle. More precisely, the POIs annotated with $M$ categories within the circle are counted and normalized to a vector $x^{(t)}_i=\{x^{(t)}_{i,1}, x^{(t)}_{i,2}, \ldots, x^{(t)}_{i,m}\}$. Thus, each stay point is transformed to store its local geographic context. Using spatial range search has the benefit of including a broader geographical context for the augmented trajectory than using a few nearest POIs, which is sensitive to outliers. 

In summary, Feature Augmentation is a crucial part of DETECT. It transforms raw GPS trajectories into scale-free, context-augmented trajectories $\{x_i\}_{i=1}^{n}$, with little loss in information but ample gain in geographical context imperative to clustering mobility behaviors.  

\subsection{Phase~\RNum{1} - Clustering with neural network} \label{subsec:phase1}
DETECT Phase~\RNum{1} operates in two steps: In the first, it constructs a continuous latent space of behaviors by learning to embed and generate context-augmented trajectories via a fully unsupervised objective. While the input is the set of augmented trajectories with arbitrary lengths, the output is fixed-length latent embedding that encodes sufficient context to facilitate accurate clustering of mobility behaviors. In the final step of this Phase, DETECT applies a clustering function on the dimension-reduced latent space. Formally, for each context augmented stay point trajectories $x_i \in \{x_i\}_{i=1}^{n}$, in the first step we adopt an RNN autoencoder~\cite{baldi2012autoencoders, lipton2015critical} with parameters $\Theta$ to learn an embedding $z_i = f_\Theta(x_i)$ in the latent space of behaviors. Finally, we apply a clustering function to output $k$ clusters, each represented by its centroid $\mu_j\in\{\mu_j\}_{j=1}^k$. 

\begin{figure}[tbp]
\centerline{\includegraphics[width=0.45\textwidth]{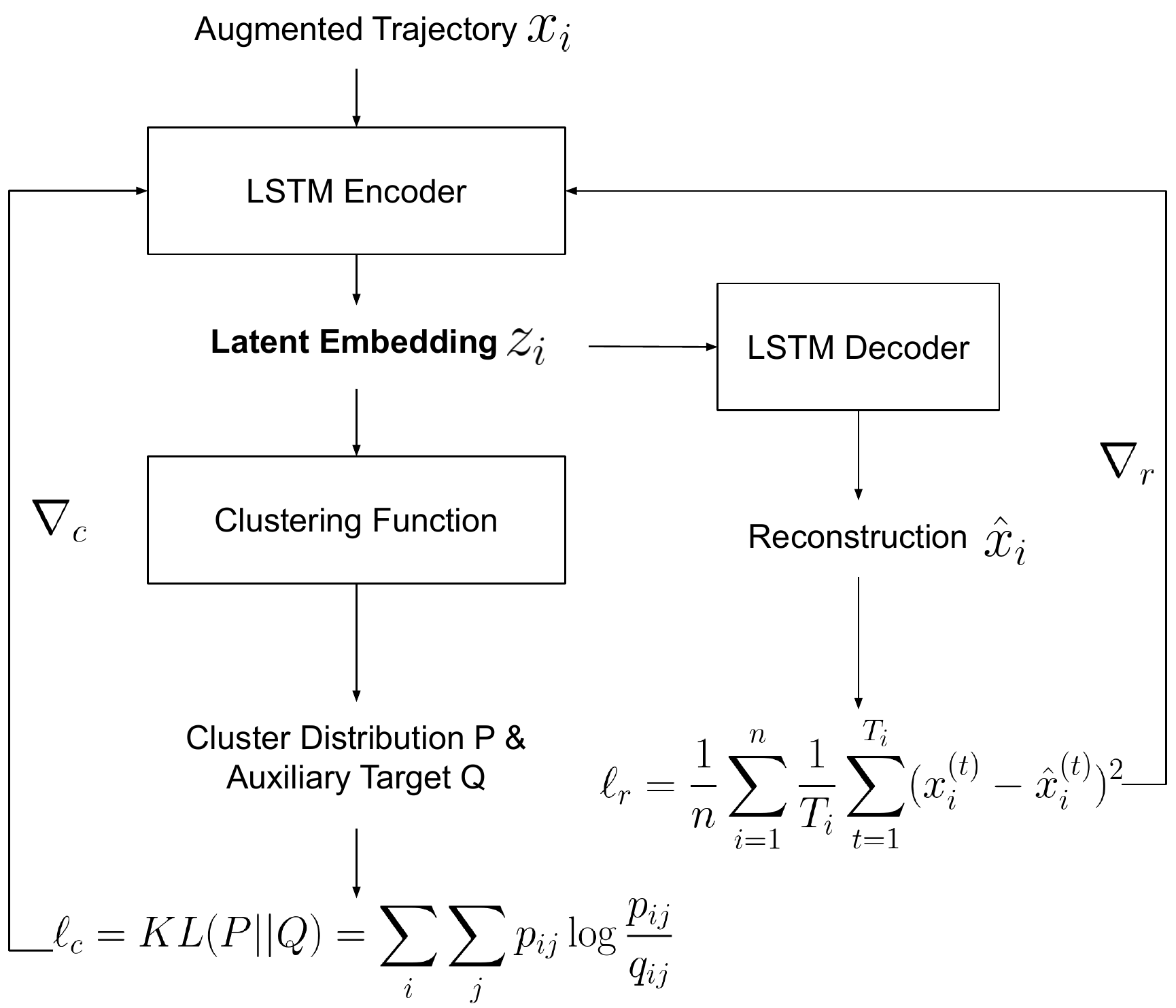}}
\caption{DETECT Phase I and Phase II: \textit{The augmented trajectories are fed to the recurrent autoencoder in Phase I. Phase I learns a hidden embedding $z$ encoding the context dynamics of the trajectories, upon which an initial cluster assignment is generated via $k$-means. Phase II jointly refines the embedding (via the encoder) and the cluster assignments.}}
\vspace{-7pt}
\label{fig:detect}
\end{figure}

The recurrent autoencoder model consists of a recurrent encoder and a recurrent decoder (see Figure~\ref{fig:detect} for an illustration). Similar to the mechanics of a general autoencoder\cite{baldi2012autoencoders}, the task of the LSTM encoder is to encode the input context augmented trajectories as a latent embedding, and then using a recurrent decoder, reconstruct the trajectories solely from the embedding. The model is trained to minimize the error of reconstruction, thus learning a representative embedding that fully captures the movement transitions and the context within a trajectory. Consider $x_i=\{x_i^{(t)}\}_{t=1}^{T_i}$ as the augmented trajectory with length $T_i$. $x_i$ is fed sequentially to the recurrent encoder comprised of several LSTM units~\cite{hochreiter1997long}. The encoder updates the hidden state $h_{enc}^{(t)}$ and other parameters with each passing unit $t$, $h_{enc}^{(t)} = \sigma(h_{enc}^{(t-1)}, x^{(t)})$, where $\sigma$ is the \textit{activation function} for the neural layer. The last hidden state $h_{enc}^{(T)}$ is called the latent embedding $z_i$, and is assumed to summarize the information necessary to represent the entire trajectory sequence. Next, the decoder tries to reconstruct the trajectory with $h_{enc}^{(T_i)}$ as its initial state, $h_{dec}^{(1)} = h_{enc}^{(T_i)}$. With $h_{dec}^{(1)}$, the decoder generates $\hat{x}^{(1)}$. The hidden states that follow are generated recursively as: $h_{dec}^{(t)} = \sigma(h_{dec}^{(t-1)}, \hat{x}^{(t-1)}),~\text{and}~ \hat{x}^{(t)} = \sigma(h_{dec}^{(t)})$. For the trajectory data, the encoder and decoder are trained together to minimize the reconstruction error:
\begin{equation}
    \ell_r = \frac{1}{n}\sum_{i=1}^{n}\frac{1}{T_i}\sum_{t=1}^{T_i}(x_i^{(t)} - \hat{x}_i^{(t)})^2
\end{equation}

The decoder can reconstruct the entire augmented trajectory from the latent embedding $z_i$, which implicitly encodes transition patterns of the geographical context. To conclude, using the neural function $z_i = f_\Theta(x_i)$, Phase~\RNum{1} maps the augmented trajectories ${x_1, x_2, \dots, x_n}$ to their corresponding fix-length embeddings ${z_1, z_2, \dots, z_n}$, before applying a clustering function (such as $k$-means on the latent space of behaviors to produce, what we call, a \textit{soft cluster assignment}.

\subsection{Phase~\RNum{2} - Joint optimization for a good clustering}\label{subsec:phase2}
A straightforward clustering over the embedded trajectories does not produce clusters tailored for mobility behavior analyses (as we show empirically in Section~\ref{sec:exp}). Therefore, in Phase II, DETECT jointly refines the embedding and the cluster assignment to improve the separability of clusters (i.e., improving both the latent embedding and the clusters of behaviors).
We achieve this using an unsupervised clustering objective, following the recent advancements in deep neural clustering~\cite{xie2016unsupervised,guo2017improved,guo2017deep}. The design of the clustering loss is based on the assumption that in the initial clusters, points that are very close to the centroid are likely to be correctly predicted/clustered, i.e., the high confidence predictions. Learning from these, the model improves the overall clustering iteratively, by aligning the low confidence counterparts. 

We develop an objective function customized to amplify the ``clustering cleanness'', i.e., minimize the similarity between clusters and maximize the similarity between points in the same cluster. In particular, the clustering refinement process iteratively minimizes the distance between the current cluster distribution $Q$ and an auxiliary target cluster distribution $P$, which is a distribution derived from high confidence predictions of $Q$. Intuitively, $Q$ describes the probability of an augmented trajectory belonging to the $k$ tentative mobility behaviors. Whereas, $P$ is generated by re-enforcing the probability of high-confidence trajectories. 

Following the work in \cite{maaten2008visualizing}, we represent the distribution of the current embedding $Q$ as the Student t-distribution on the current cluster centers. Given the centroid of the $j^{\text{th}}$ cluster $\mu_j$, and the latent embedding of the $i^{\text{th}}$ trajectory $z_i$, we calculate the current distribution as
\begin{equation}
\label{eq:q_dist}
    q_{ij} = \frac{(1+||z_i-\mu_j||^2)^{-1}}{\sum_{j'}{(1+||z_i-\mu_{j'}||^2)^{-1}}}
\end{equation}
where $q_{ij}$ is the probability of assigning $z_i$ to $\mu_j$, or read as the probability of assigning trajectory $x_i$ to the $j^{\text{th}}$ tentative mobility behavior.
In the current distribution, the low-confidence points are assumed to be assigned to poor clusters, and hence in the need of refinement to better clustering cleanness. 

We derive an auxiliary distribution made up of the high confidence assignments of the current distribution. The goal of this self-training target distribution is to 1) emphasize data points assigned with high confidence, and (2) normalize loss contribution to prevent the sizes of clusters from negatively impacting the latent embeddings. Equation~\eqref{eq:p_dist} defines our target distribution $P$, where $p_i$ is computed by first raising $q_i$ to the second power and then normalizing by the frequency per cluster. The second power of probabilities places more weight on the instances near the centroids. The division of $\sum_{i'}{q_{i'j}}$ normalizes the different cluster sizes, making the model robust to biased classes in the data. 
\begin{equation}
\label{eq:p_dist}
    p_{ij}=\frac{q_{ij}^2 / \sum_{i'}{q_{i'j}}}{\sum_{j'}({q_{ij'}^2/\sum_{i'}{q_{i'j'}})}}
\end{equation}

To measure the distance between the distributions $P$ and $Q$ (as defined above), we use the K-L divergence, a widely known distribution-wise asymmetric distance measure. The clustering oriented loss is defined as, 
\begin{equation}
\label{eq:KL}
    \ell_c = KL(P||Q)=\sum_i\sum_j p_{ij}\log\frac{p_{ij}}{q_{ij}}
\end{equation}
Lastly, as the most computationally intensive step, we iteratively minimize the distance between the soft clustering assignment $Q$ and the auxiliary distribution $P$ by jointly training the recurrent encoder (in turn, the latent embedding), and the cluster assignment. We train our model using the Stochastic Gradient Descent optimization minimizing the clustering loss defined as a K-L divergence. Updates are made in an iterative manner to the latent trajectory embedding---through training on its own high confidence clustering assignments and refining cluster centroids---resulting in distinct mobility behavior boundaries. For completeness, we present the computation of the gradients $\frac{\partial \ell}{\partial z_i}$ and $\frac{\partial \ell}{\partial \mu_j}$ in equation~\ref{eq:gradient}, and leave the derivation of the gradients propagated backward to the recurrent encoder (i.e. $\frac{\partial \ell}{\partial \Theta}$) as an exercise.
\begin{equation}
\label{eq:gradient}
\begin{split}
 &\frac{\partial \ell}{\partial z_i} = 2\sum_{j=1}^{k}(z_i-\mu_j)(p_{ij}-q_{ij})(1+\left \| z_i-\mu_j \right \|^2)^{-1} \\ 
 &\frac{\partial \ell}{\partial \mu_j} = 2\sum_{i=1}^{n}(z_i-\mu_j)(q_{ij}-p_{ij})(1+\left \| z_i-\mu_j \right \|^2)^{-1} 
\end{split}
\end{equation}

The final output of DETECT is an encoder that is finely tuned to learn a data representation specialized for clustering without groundtruth cluster membership labels.
\textit{}\section{Experimental Evaluation \note{(3 page)}}
\label{sec:exp}
In this section, we quantitatively and qualitatively evaluate DETECT for trajectory clustering with extensive experiments on two real-world trajectory datasets.

\subsection{Experimental Settings} 
\label{sec:dataset}
\textbf{Datasets.} We utilize two datasets for the evaluation of our proposed approach: the GeoLife dataset~\cite{zheng2009mining} and the DMCL dataset~\cite{dmcl}. The \textit{GeoLife} dataset consists of 17,621 trajectories generated by 182 users from April 2007 to August 2012. We validate our approach against the ground-truth. we extract a subset of the data comprised of 601 trajectories from 11 users. This is a common methodology for mobility analysis and its applications \cite{zhou2018trajectory, cui2018personalized, chang2018content}. Furthermore, we retrieve POI information from the PKU Open Research Data~\cite{pkupoi}. This dataset contains over 14,000 POIs in Beijing, falling into 22 major categories including ``education'', ``transportation'', ``company'' and ``shopping''. The \textit{DMCL} dataset consists of trajectories in Illinois, United States. It contains 90 complete trajectories generated by two users over six months, and is used as a whole. The POI data is scraped from OpenStreetMap (OSM)~\cite{osm}. It contains 30,401 POIs in Illinois, subject to 9 major categories such as ``public'' and ``accommodation''. Table \ref{tab:preprocess} gives some statistics on the duration and lengths of the trajectories in both datasets.

\begin{table}
\centering
\caption{Stats of datasets}
\label{tab:preprocess}
\begin{tabular}{cccccc}
\toprule
Dataset & stats                 & min & max & mean & std. \\
\midrule
GeoLife & Duration (min)    & 0.91  & 1177.96 & 192.55 & 257.62 \\
        & Length (km) & 0.01  & 11.68   & 2.36   & 2.15   \\
DMCL    & Duration (min)    & 15.45 & 651.2   & 314.21 & 234.42 \\
        & Length (km) & 0.004 & 38.9    & 11.84  & 17.65\\
\bottomrule
\end{tabular}
\end{table}

\noindent\textbf{Data Preparation and Ground-truth.}
The ground-truth is prepared by manually labelling the datasets. It is performed by an expert, through a meticulous process, bereft of any knowledge on the clustering membership labels. More precisely, labels are generated in an iterative manner by first visualizing the trajectory by georeferencing all its GPS points on the map using Mapbox Gl\footnote{\url{https://github.com/mapbox/mapboxgl-jupyter}}. Subsequently, the duration of time between any two consecutive check-ins and the type of surrounding buildings are studied via a Google Maps plugin, before making the judgement on its mobility behavior.  In GeoLife dataset, six mobility behaviors were identified as the ground-truth classes: ``campus activities'', ``hangouts'', ``dining activities'', ``healthcare activities'', ``working commutes'', ``studying commutes''. While in the DMCL dataset, four mobility behaviors were identified: ``studying commutes'', ``residential activities'', ``campus activities'', ``hangouts''.  The labelling process consumed 60 hours of labor and the generated dataset is publicly released  here\footnote{\url{https://tinyurl.com/y5a3r3oy}}.\\

\noindent\textbf{Compared Approaches.} We evaluate the following approaches:
\begin{itemize}[leftmargin=\parindent,align=left,labelwidth=\parindent,labelsep=0pt]
    \item \textbf{KM-DBA~\cite{petitjean2011global,Petitjean2014ICDM}:} DBA stands for Dynamic Time Warping Barycenter Averaging. It utilizes DTW as its distance measure for $k$-means clustering on sequential data, before performing a sophisticated averaging step. We set parameter $k=6~(4)$ for the GeoLife (DCML) dataset, respectively.
    \item \textbf{DB-LCSS~\cite{morris2009learning}:} DB-LCSS uses DBSCAN, a density-based clustering approach, with LCSS as its distance measure between raw trajectories. For GeoLife (DCML, respectively) we set the common sequence threshold as $0.15~(1.5\text{e-}6)$ for LCSS, and $\epsilon=0.03~(1\text{e-}6)$ and $minPts=18~(2)$ as the neighborhood thresholds in DBSCAN.
    \item \textbf{SSPD-HCA~\cite{besse2016review}:} Symmetrized Segment-Path Distance is a \emph{shape-based} distance metric particularly suited to measuring similarity between location trajectories. It utilizes an agglomerative hierarchical clustering procedure with the \emph{Ward's} criterion for choosing the pair of clusters to merge at each step.
    \item \textbf{KM-DBA*, DB-LCSS*:} For a more interesting comparison, we adapt KM-DBA and DB-LCSS methods to work on context augmented trajectories.
    \item \textbf{RNN-AE~\cite{yao2018learning,yao2017trajectory}:} We label as RNN-AE the simple process of training a recurrent autoencoder on the segmented \textit{raw trajectories} and then clustering the learned embedding using $k$-means.
    \item \textbf{DETECT Phase~\RNum{1}, DETECT:} We evaluate two variations of DETECT, the first one, termed "DETECT Phase~\RNum{1}" only includes the first phase, while the second variant, termed "DETECT" includes both phases \RNum{1} and \RNum{2}. RNN-AE is identical to DETECT Phase~\RNum{1} in all aspects except that it is trained on raw trajectory data.
\end{itemize}
Finally, we note that in the interest of space we omit the comparison of clustering with 1) HU distance \cite{hu2007semantic} and 2) PCA decomposition. HU distance is computed as the average Euclidean distance between points on two trajectories. The PCA distance
is similar to HU but works in a lower dimensional space via PCA
decomposition. Both methods are inferior to the above baselines for trajectory clustering~\cite{morris2009learning}.\\

\noindent\textbf{Evaluation metrics.}
We evaluate the extent to which cluster labels match externally supplied class labels according to four well-established \textit{external} metrics: 1) Rand Index (RI) measures the simple accuracy, i.e., the percentage of correct prediction of clusters. 2) Mutual Information (MI) measures the mutual dependency between the clustering result and the ground-truth, i.e., how much information can one infer from the other. Zero mutual information indicates the clustering labels that are independent from the ground-truth classes. 3) Purity measures how pure are the clustering results, i.e., whether the trajectories in the same cluster belong to the same ground-truth class. 4) Fowlkes-Mallows Index (FMI) measures the geometric mean of the pairwise precision and recall, which is robust to noises. \\

\noindent\textbf{Training.} We implement our approaches on a computer with an Intel Core i7-8850H CPU, a 16 GB RAM and an NVIDIA GeForce GTX 1080 GPU. 
We implement the KM-DBA using tslearn~\cite{tslearn}, and DBSCAN clustering using Sklearn with LCSS distance~\footnote{\url{https://github.com/maikol-solis/trajectory_distance}}.
The proposed deep embedded neural network is built using Keras~\cite{chollet2015keras} with Tensorflow~\cite{abadi2016tensorflow}.

\subsection{Performance Comparison with Baselines}
We quantitatively evaluate the clustering quality of DETECT against all baselines on the GeoLife Dataset. The improvements of DETECT over the compared approaches all passed the paired t-tests with significance value $p<0.03$. The results are depicted in Table~\ref{tab:compareall}. DETECT clearly outperforms all compared approaches. The relative performance against the baseline approaches varies across metrics; ranging from at least a 41\% improvement (in FMI against KM-DBA) and up to 252\% improvement (in RI against DB-LCSS). Even the primitive neural approach RNN-AE competes in performance with the alignment based methods, demonstrating the advantage of a neural approach in modeling the transitions within raw GPS trajectories. Moreover, DETECT Phase~\RNum{1} produces significant quality improvements over highly customized distance-based metrics such as SSPD, confirming that the autoencoder trained on context augmented trajectories can accurately model context transitions in the latent space of behaviors.

\begin{table}
\centering
\caption{Clustering performance of all approaches.}
\label{tab:compareall}
\begin{tabular}{ccccc}
\toprule
Method            & RI      & MI & Purity          & FMI             \\ \midrule
KM-DBA       & 0.33         & 0.64            & 0.58          & 0.58          \\
DB-LCSS       & 0.22          & 0.55             & 0.51          & 0.56          \\
RNN-AE       & 0.39          & 0.46             & 0.56          & 0.53          \\
SSPD-HCA    & 0.52    &    0.93         &   0.66      &   0.67          \\
KM-DBA*       & 0.51          & 0.91             & 0.74          & 0.63          \\
DB-LCSS*       & 0.50          & 0.95             & 0.64          & 0.66          \\
DETECT Phase I    & 0.65          & 1.06             & 0.84           & 0.73          \\
DETECT & \textbf{0.76} & \textbf{1.26}    & \textbf{0.89} & \textbf{0.81} \\ 
\midrule
Impr. over KM-DBA    & 132\% & 98\% & 54\% & 41\% \\
Impr. over DB-LCSS  & 252\% & 131\% & 74\% & 46\% \\
\bottomrule
\end{tabular}
\end{table}

\subsection{Ablation Study}\label{subsec:ablation}
In order to understand the influence of feature augmentation, neural embedding, and cluster refinement individually, we conduct an \textit{ablation study} by isolating the effects of each procedure.
 
\subsubsection{Benefit of stay point detection and geographical augmentation}
On the GeoLife dataset, Table~\ref{tab:preprocess} presents the clustering quality of KM-DBA on trajectories with progressive levels of feature augmentation: 1) \textit{raw trajectory} is the simple sequence of GPS points; 2) \textit{stay points only trajectory} is the sequence of spatiotemporal points extracted from the raw trajectory by Fast-SPD; 3) \textit{geographical only trajectory} is the sequence of geographical vectors generated at each point in the raw trajectory, rather than at stay points; and 4) \textit{fully augmented trajectory} is the sequence of geographical vectors generated at each stay point extracted from the raw trajectory. 

It is clear that the methods using fully augmented trajectories outperforms methods with a weaker degree of augmentation. The stay points only trajectory data are comparable in performance to raw trajectory input because both do not incorporate geographical context that is essential to learning a rich embedding in the latent space of mobility behaviors. As a result, the geographical only trajectory data has better accuracy than both. However, when context augmentation is combined with stay point extraction (i.e. fully augmented trajectories), the two procedures complement each other. Fast-SPD discards the irrelevant geographical context features of GPS points that lie in-between the critical stay points of a trajectory. Thus, achieving a result that is better than the sum of its parts.

\begin{table}
\centering
\caption{Performance with varying level of augmentation.}
\label{tab:preprocess}
\begin{tabular}{ccccc}
\toprule
Data Type            & RI      & MI & Purity          & FMI             \\ \midrule
Raw trajectory       & 0.33          & 0.64             & 0.58          & 0.58          \\
Stay point only       & 0.30          & 0.68             & 0.60          & 0.57          \\
Geographical only    & 0.44          & 0.85             & 0.69          & 0.59          \\
Augmented trajectory & \textbf{0.52} & \textbf{0.93}    & \textbf{0.75} & \textbf{0.63} \\ 
\bottomrule
\end{tabular}
\vspace{-7pt}
\end{table}

\subsubsection{Benefit of neural network and cluster refinement} 
\label{sec:cluster_comparsion}
\begin{figure}
    \centering
    \begin{subfigure}[b]{0.45\columnwidth}
        \centering
        \includegraphics[width=\columnwidth]{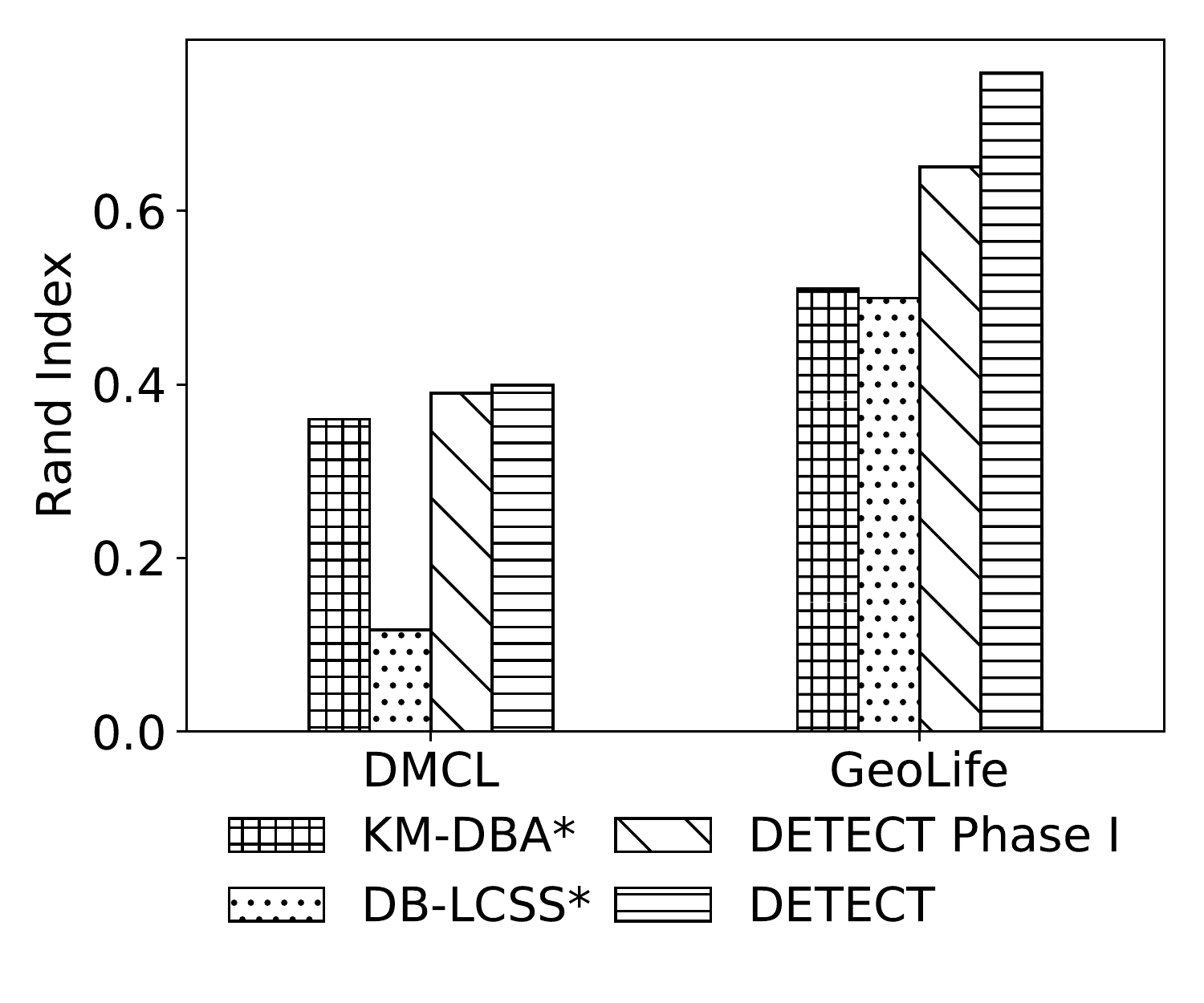}
        \caption{Rand Index}
    \end{subfigure}
    \begin{subfigure}[b]{0.45\columnwidth}
    \centering
        \includegraphics[width=\columnwidth]{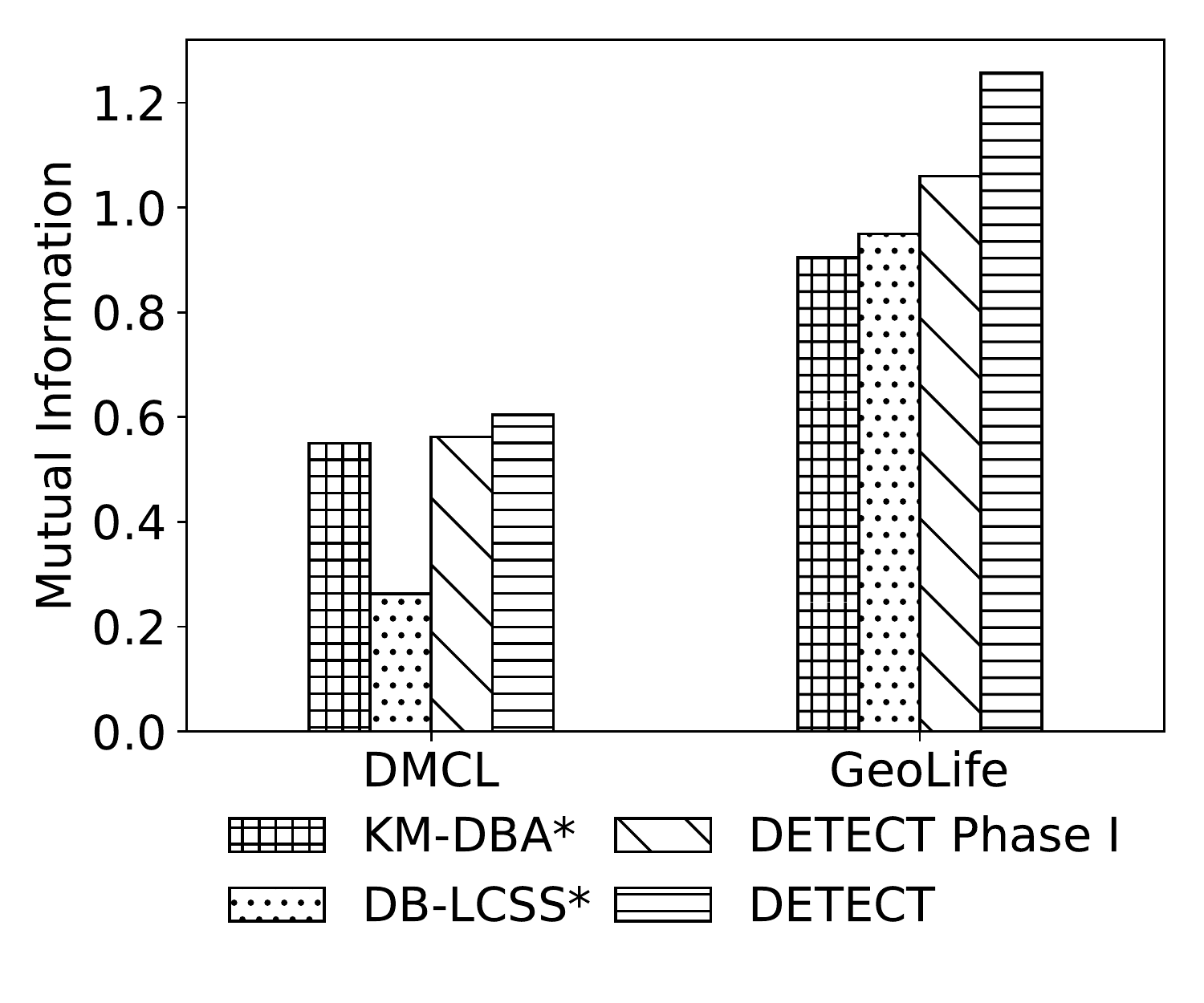}
        \caption{Mutual Information}
    \end{subfigure}
    \begin{subfigure}[b]{0.45\columnwidth}
    \centering
    \includegraphics[width=\columnwidth]{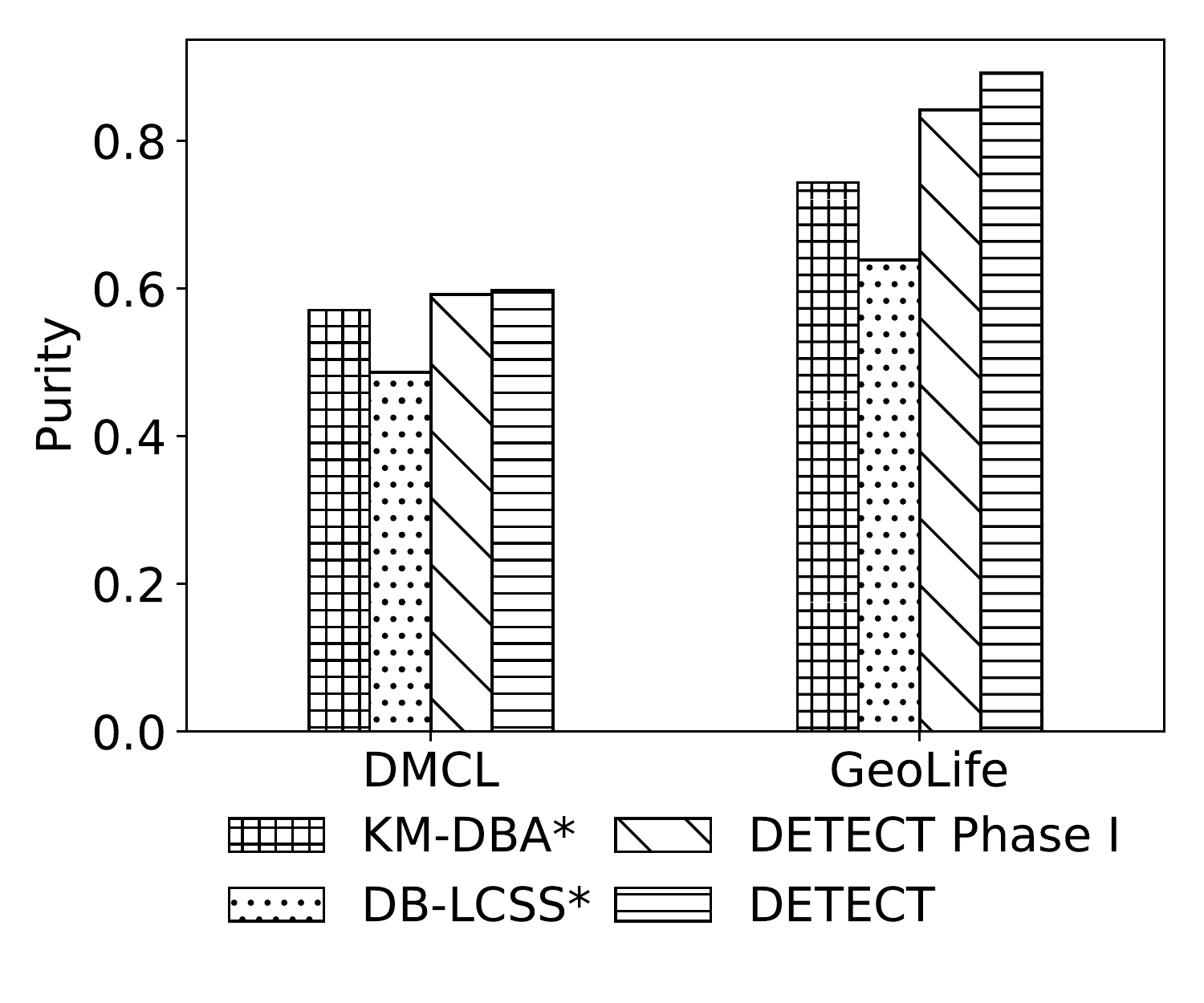}
        \caption{Purity}
    \end{subfigure}
    \begin{subfigure}[b]{0.45\columnwidth}
    \centering
    \includegraphics[width=\columnwidth]{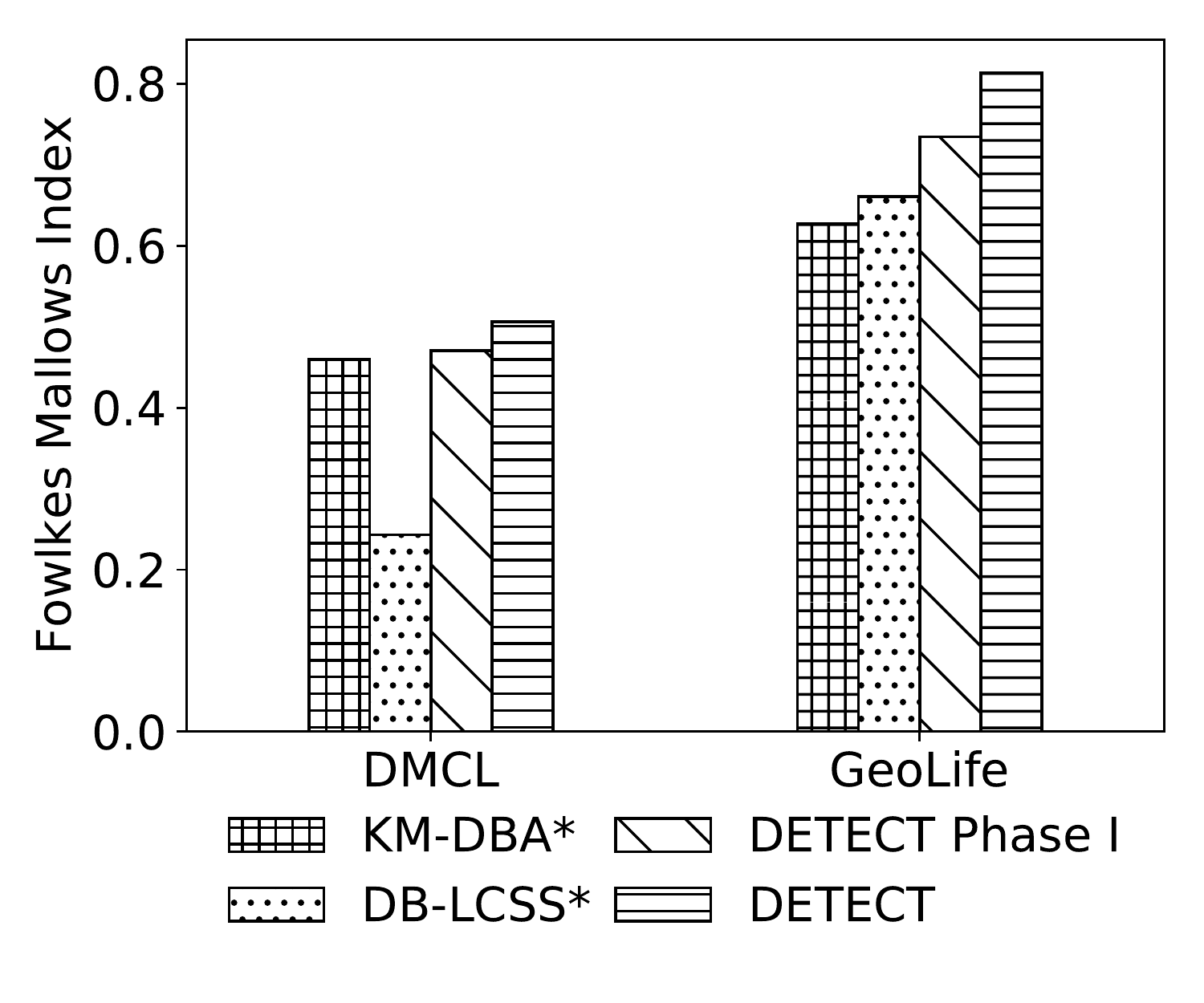}
        \caption{Fowlkes Mallows Index}
    \end{subfigure}
    \caption{Ablation study on compared approaches.\\ DETECT outperforms the adapted baselines with feature augmentation (marked with *) on all metrics and for both datasets (relatively little improvements are seen in DMCL, since it spans only two users with limited mobility variation).}
    \label{fig:clustering}
\end{figure}

There is no previous study on clustering mobility behaviors that uses a neural network to model context transitions in trajectories or one that iteratively refines behavior clusters. Therefore, we are interested in evaluating how the deep neural architecture can benefit mobility behavior analyses. To isolate the contribution of the neural network, we set the inputs of all approaches in this experiment as context augmented trajectories. Accordingly, we compare adapted baselines KM-DBA* and DB-LCSS*, against neural approaches DETECT Phase~\RNum{1} and DETECT.  Figure~\ref{fig:clustering} presents the results on both datasets.  
The improvements of DETECT are more significant in the GeoLife dataset than in the DMCL dataset. Training on a larger dataset (GeoLife) produces a latent embedding that better captures the transitions in a trajectory. But more importantly, since DMCL is comprised of only two users, it contains trajectories with limited variation in mobility behaviors (e.g., a user generally visits the same supermarket), accordingly the benefit of our approach is relatively small. Based on the results, it is evident that DETECT Phase~\RNum{1} learns an expressive latent embedding in the trajectories, while other baselines rely on alignments between augmented trajectories and fail to capture the dynamics of the data. 

The improvement by clustering refinement is significant as illustrated in the difference in clustering quality between DETECT Phase~\RNum{1} and DETECT. It proves that the customized clustering oriented loss offers large benefits in clustering accuracy by helping separability between mobility behaviors.

\subsection{Parameter study}
\label{sec:parameter_study}
Below we discuss the effects of different parameter settings of DETECT on the reconstruction and clustering performance. 

\subsubsection{Effect of the number of clusters}
We illustrate the effect of varying $k$ on the clusters in Figure~\ref{fig:para:k}. The experiment is conducted on the GeoLife dataset with six ground-truth classes.
Both DETECT Phase~\RNum{1} and DETECT reach their best FMI at around six. 
This aligns with the intuition that the model performs best as the number of clusters approaches the number of ground-truth classes. 
In practice however, supervision is not available for hyperparameter cross-validation, hence alternate methods are employed to determine the clustering cardinality. Most common methods include manually determining the number of clusters via low dimensional visualizations or tuning $k$ through Elbow method~\cite{bholowalia2014ebk} using \textit{internal} metrics~\cite{liu2010understanding}. 

\subsubsection{Effect of varying thresholds in Fast-SPD and geographical augmentation}
Parameters $\rho_s$ and $\rho_t$ in Fast-SPD control the magnitude of stay points extracted. Usually the algorithm is robust to reasonable practical values, e.g. ~1km for $\rho_s$ and ~20 min for $\rho_t$. But too small or too large thresholds tend or cause over-extraction or nothing extracted from the trajectories. Similar robustness if observed with the buffer size $r_{poi}$ of geographical context augmentation. However, a small value generates a context vector very sensitive to outliers, while a large value includes POIs that are less likely to be important , reducing the variation between stay points, and obscuring the transitions of contexts. In this paper, we set $r_{poi} = 1 km$ but values between 0.5km to 1.5km yield almost as good a result. Overall, robustness is an important property of the feature augmentation part of DETECT.

\subsubsection{Effect of the latent embedding dimension}
Table~\ref{tab:tune_d} demonstrates the effect of latent embedding dimension $d$ on the reconstruction error in Phase I. Given a small number of hidden dimensions, the model is incapable of learning an expressive-enough latent embedding. Whereas, if the number of dimensions grows too large, the model easily overfits the training data. Overall, for a value of $d$ between 50 and 100, the clustering performance remains good.

\begin{table}[ht]
\centering
\caption{Mean and Standard Deviation of MAE for DETECT Phase I with varying latent embedding dimension $d$}
\label{tab:tune_d}
\begin{tabular}{lcccc}
\toprule
$\mathbf{d}$  & 16       & 32     & 64       & 128       \\ \midrule
\textbf{mean ($\times10^{-3}$)} & 5.6   & 4.9 & 4.3 & 4.5  \\
\textbf{std  ($\times10^{-3}$)}  & 0.61 & 0.2 & 0.06 & 0.23 \\ 
\bottomrule
\end{tabular}
\end{table}



\subsubsection{Effect of the learning rate and training epochs}
\begin{figure}
    \centering
    \begin{subfigure}[b]{0.48\columnwidth}
        \centering
        \includegraphics[width=\columnwidth]{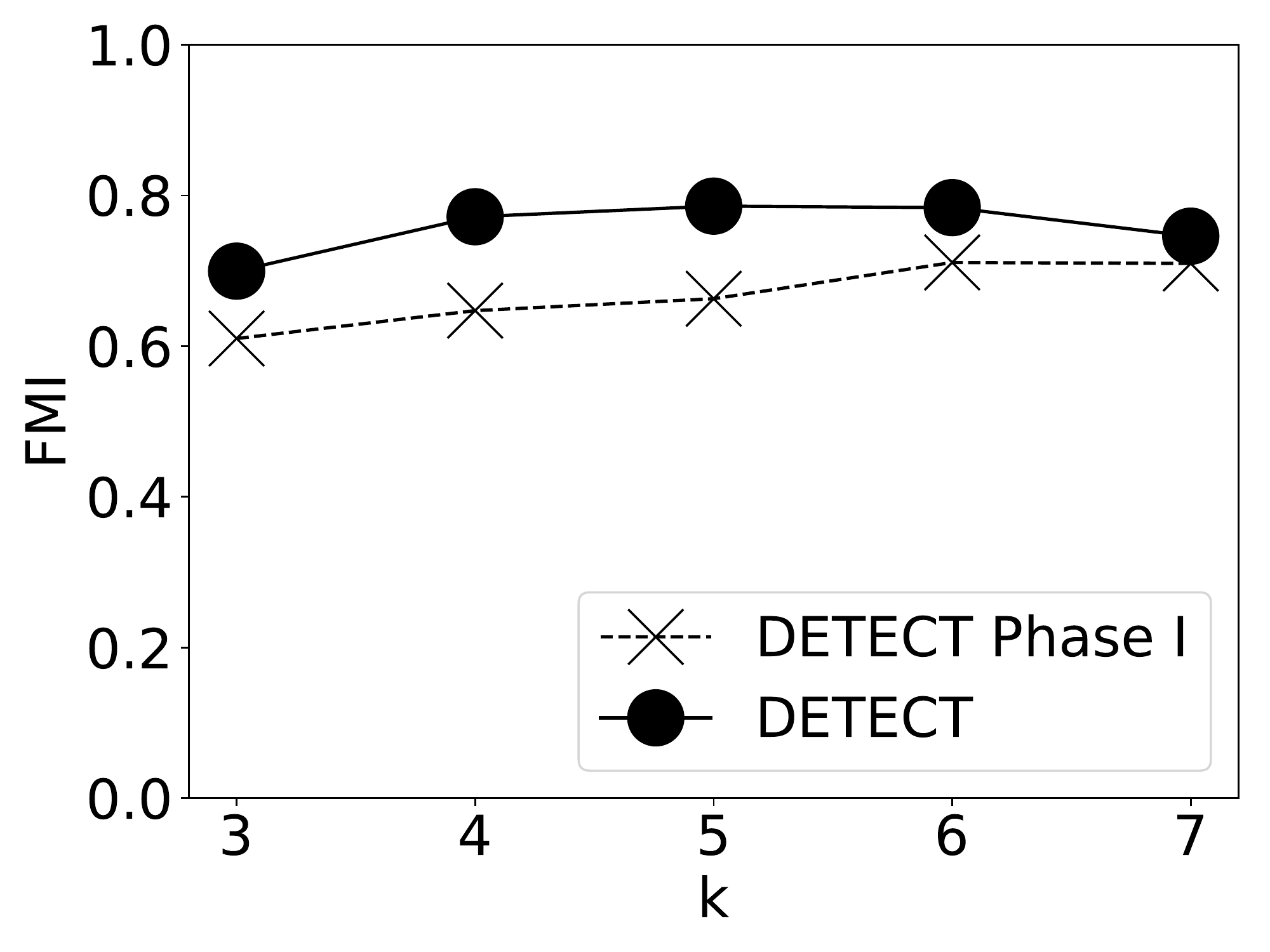}
        \caption{}
        \label{fig:para:k}
    \end{subfigure}
    \begin{subfigure}[b]{0.48\columnwidth}
    \centering
         \includegraphics[width=\columnwidth]{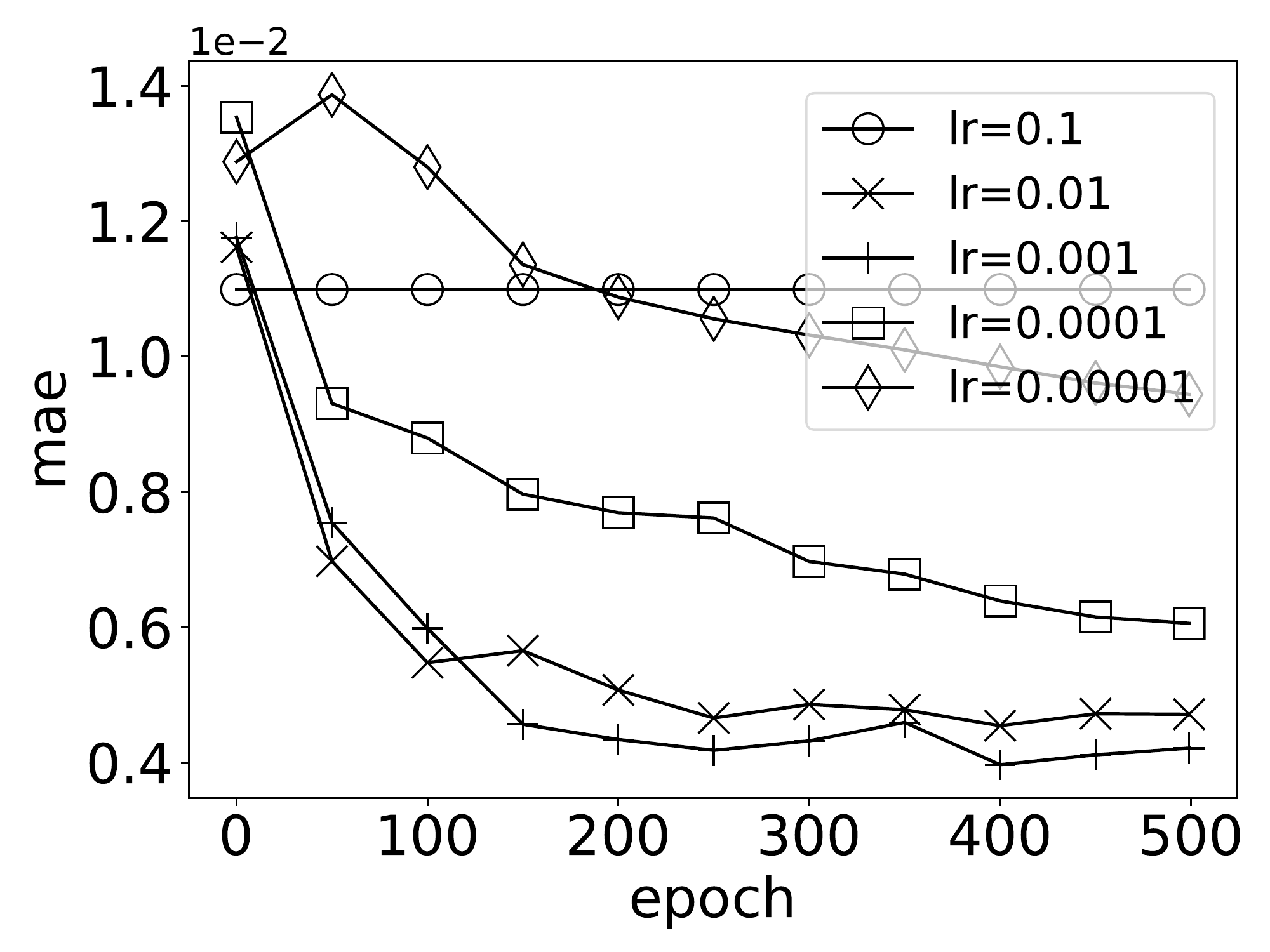}
        \caption{}
        \label{fig:para:lr}
    \end{subfigure}
    \caption{Effect of DETECT parameters. (a) Effect of the number of clusters $k$. (b) Effect of the learning rate $lr$}
    \label{fig:para}
\end{figure}

In Figure~\ref{fig:para:lr}, we compare the training curves with different learning rates. Given a large learning rate, i.e., $lr = 0.1$ the model incurs a proportionally large reconstruction error. In contrast, when learning rate is too small, e.g., $lr = 10^{-5}$, the model takes long to converge. However, for a reasonable value of the parameter, the unsupervised training loss converges fast after the first few epochs. We also remark that the model does not overfit the dataset too readily. We notice (not shown here) an increase in standard deviation of MAE only when the model is trained for more than 1500 epochs.

\subsection{Qualitative Evaluation}
\label{sec:viz}
\begin{figure}
    \centering
    \begin{subfigure}[b]{0.48\columnwidth}
    \centering
    \includegraphics[width=\columnwidth]{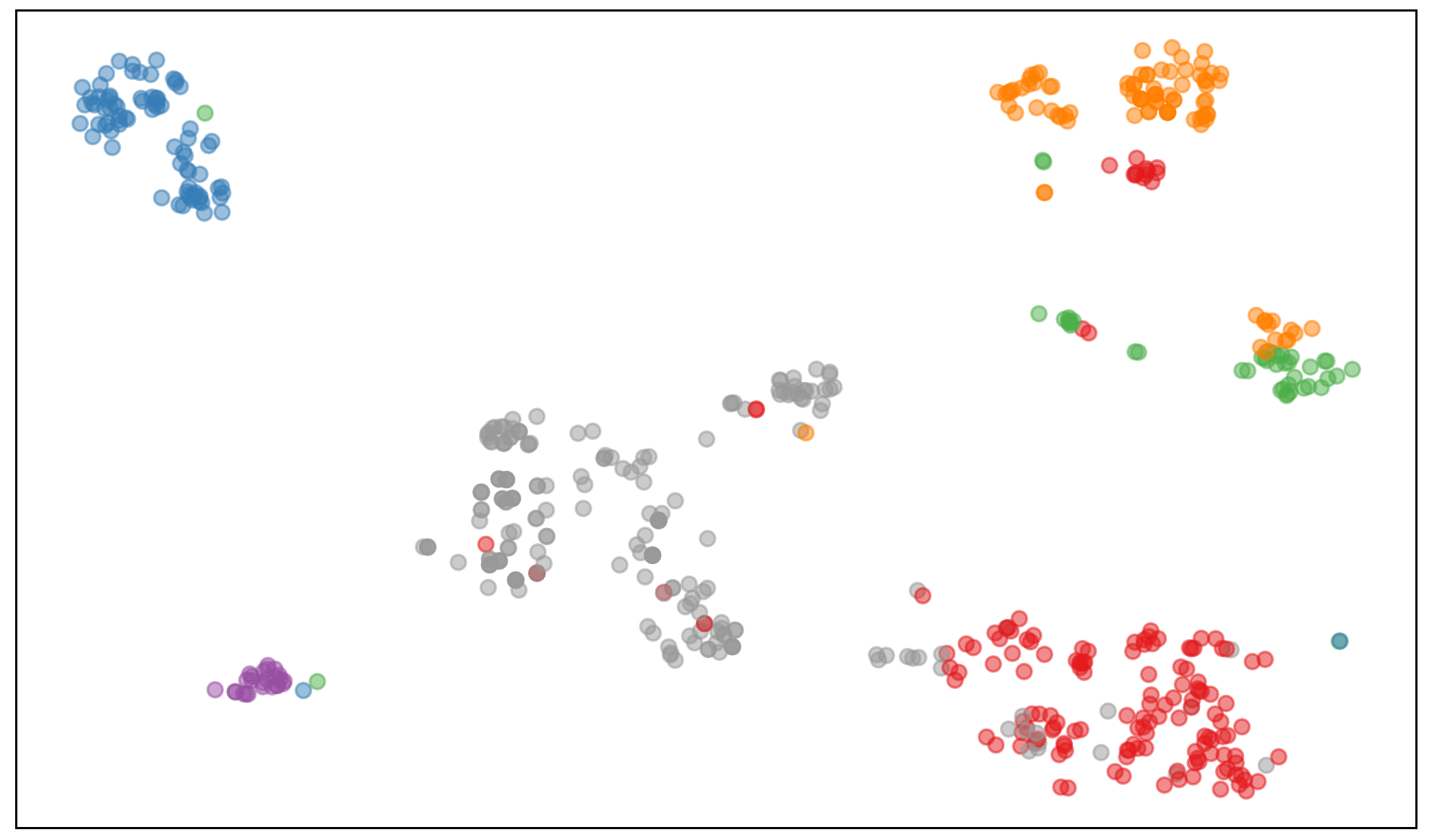}
        \caption{DETECT}
        \label{fig:tsne:phase2}
        \end{subfigure}
    \begin{subfigure}[b]{0.48\columnwidth}
        \centering
        \includegraphics[width=\columnwidth]{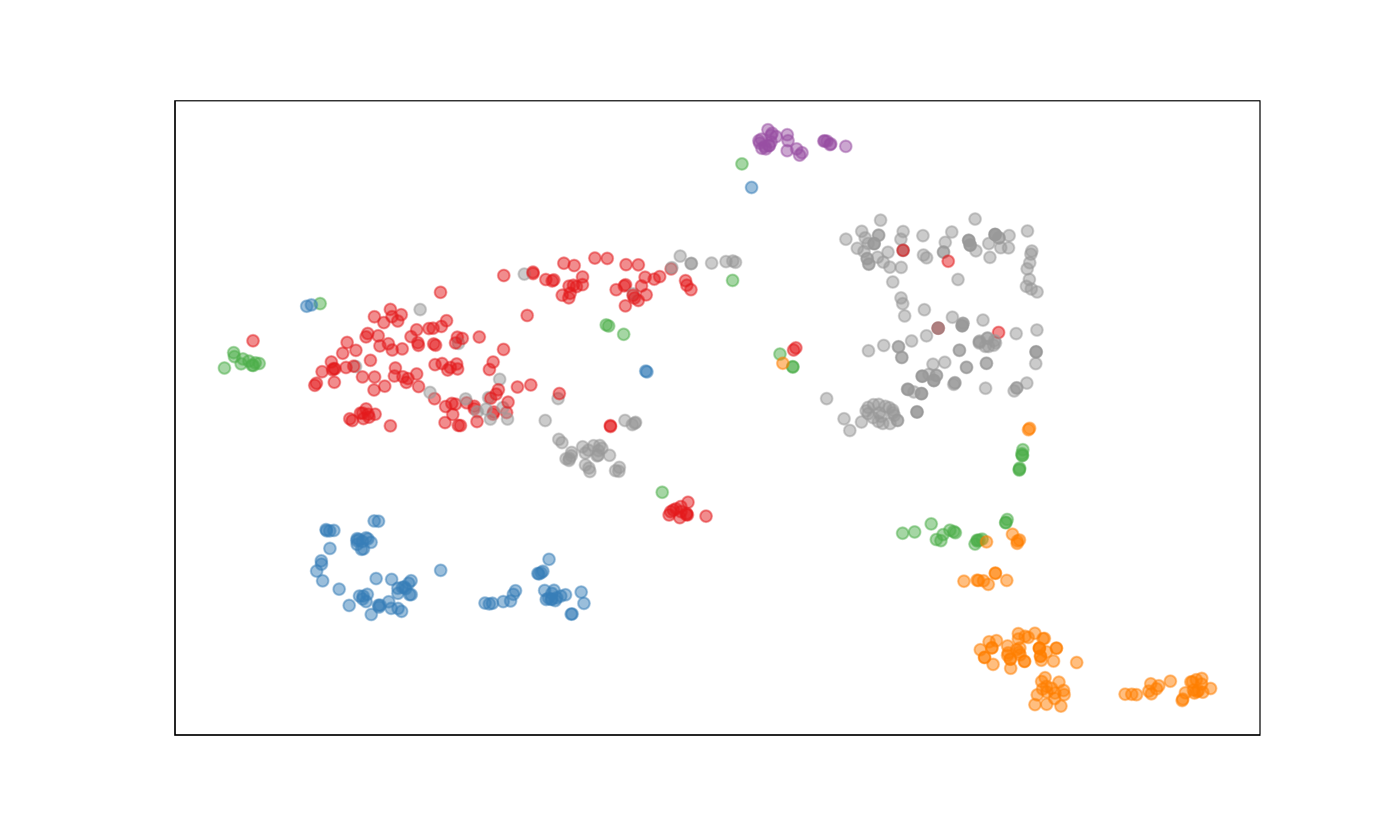}
        \caption{DETECT Phase I Only}
        \label{fig:tsne:phase1}
    \end{subfigure}
    \caption{Clustering results visualization of ``DETECT'' and ``DETECT Phase I'' using t-SNE with perplexity 40}
    \label{fig:tsne}
\end{figure}

\textbf{Visualization study.} To further understand the learned latent embedding of DETECT, we generate a series of visualizations.
Figure~\ref{fig:tsne} shows the two-dimensional  t-SNE~\cite{maaten2008visualizing} plot over DETECT embeddings on the GeoLife dataset.
Each point is colored based on the corresponding ground-truth class.
DETECT generate well-formed clusters (Figure~\ref{fig:tsne:phase2}) with most points in the same class grouped into the same cluster while being well separated from others. A close inspection also reveals that the clusters in Figure~\ref{fig:tsne:phase2} are ``cleaner'' than the ones in Figure~\ref{fig:tsne:phase1}, demonstrating the effectiveness of cluster refinement in Phase II, wherein the embedding and the clustering assignment are jointly optimized to gain better differentiation in mobility behaviors. 

\begin{figure}[t]
    \centering
    \begin{subfigure}[b]{0.7\columnwidth}
        \centering
        \includegraphics[width = \columnwidth]{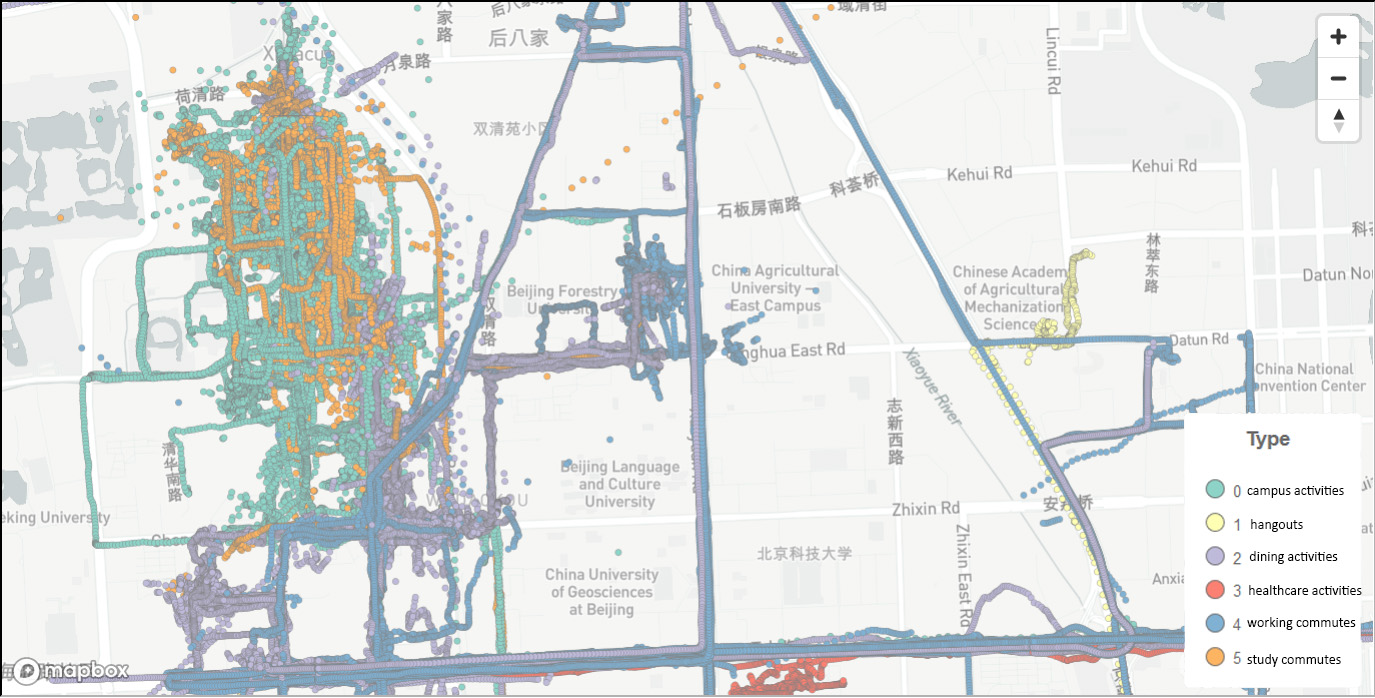}
        \caption{Trajectories with predicted class}
        \label{fig:map1}
    \end{subfigure}
    \begin{subfigure}[b]{0.7\columnwidth}
    \centering
        \includegraphics[width =\columnwidth]{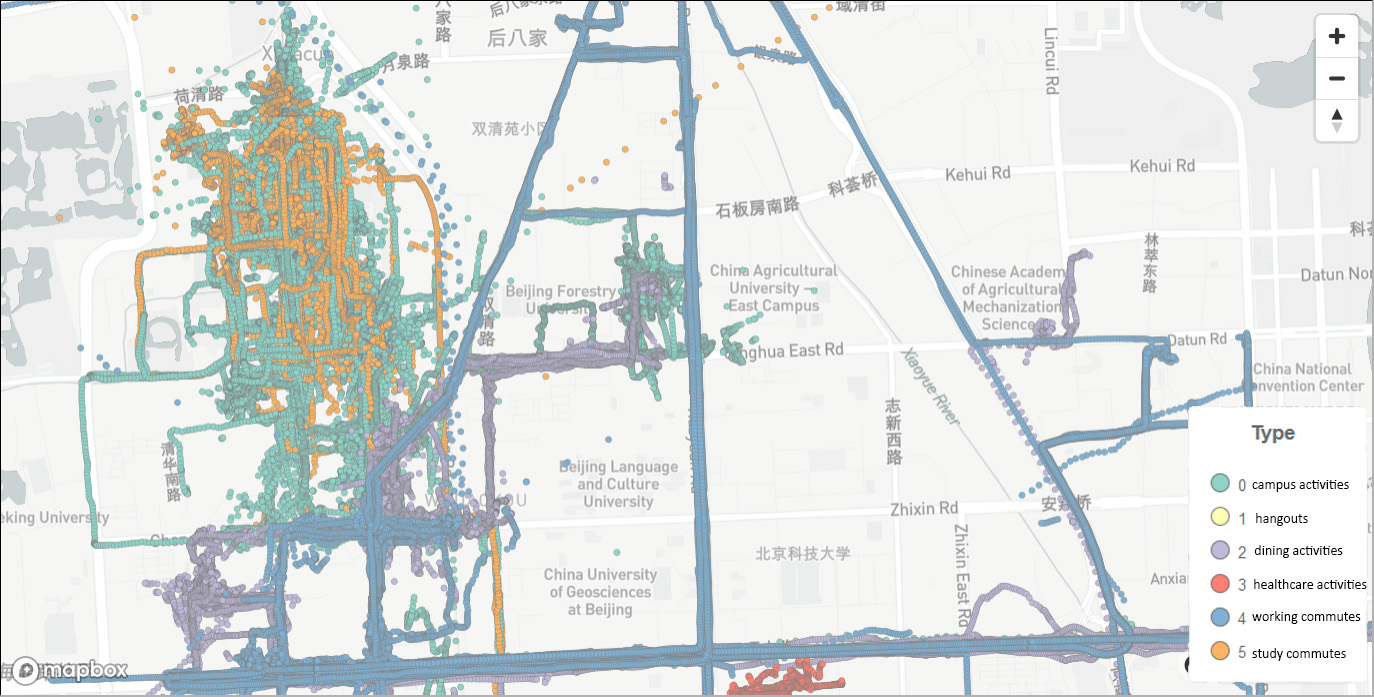}
        \caption{Trajectories with ground-truth class}
        \label{fig:map2}
    \end{subfigure}
    \caption{Projecting raw trajectories clusters onto the map. (a) Colored based on predicted cluster labels (b) Colored based on ground-truth classes}
    \label{fig:map}
    \vspace{-7pt}
\end{figure}

Figure~\ref{fig:map1} visualizes the trajectories with the predicted classes and figure~\ref{fig:map2} depicts the ground-truth classes in the GeoLife dataset. The individual predicted classes are labeled using the same color as their corresponding ground-truth classes. It is evident that the predicted classes generally match the ground-truth, even though each class contains trajectories with various shapes and lengths. This is a testament to DETECT's capabilities in clustering trajectories of widely different spatial and temporal range of movement.\\

\vspace{-10pt}
\subsection{Scalability}
We evaluate DETECT on the entire GeoLife dataset comprised of the 17,621 trajectories. Since the full dataset is unlabelled, we utilize four \textit{internal} validation measures to understand the compactness, the connectedness and the separation of the cluster partitions. Namely, the \textit{Silhouette score}, which ranges from -1 to +1, where a high value indicates that the object is well matched to its own cluster and poorly matched to neighboring clusters. \textit{Dunn Index} tries to maximise intercluster distances whilst minimising intracluster distances. Thus, large values of Dunn Index correspond to good clusters. The \textit{Within-Like} criterion, captures the intracluster variance, accordingly a small value indicates compact clusters. Likewise, the Between-Like criterion captures intercuster variance, hence a large value indicates a good separation between different clusters. These metrics are customized to support context-augmented trajectories and are formally defined in~\cite{besse2016review}. Note that internal metrics are not suitable for comparison between clustering approaches that utilize different distance measures (e.g. DETECT vs DB-LCSS). 

\begin{figure}
    \centering
    \begin{subfigure}[b]{0.47\columnwidth}
        \centering
        \includegraphics[width=\columnwidth]{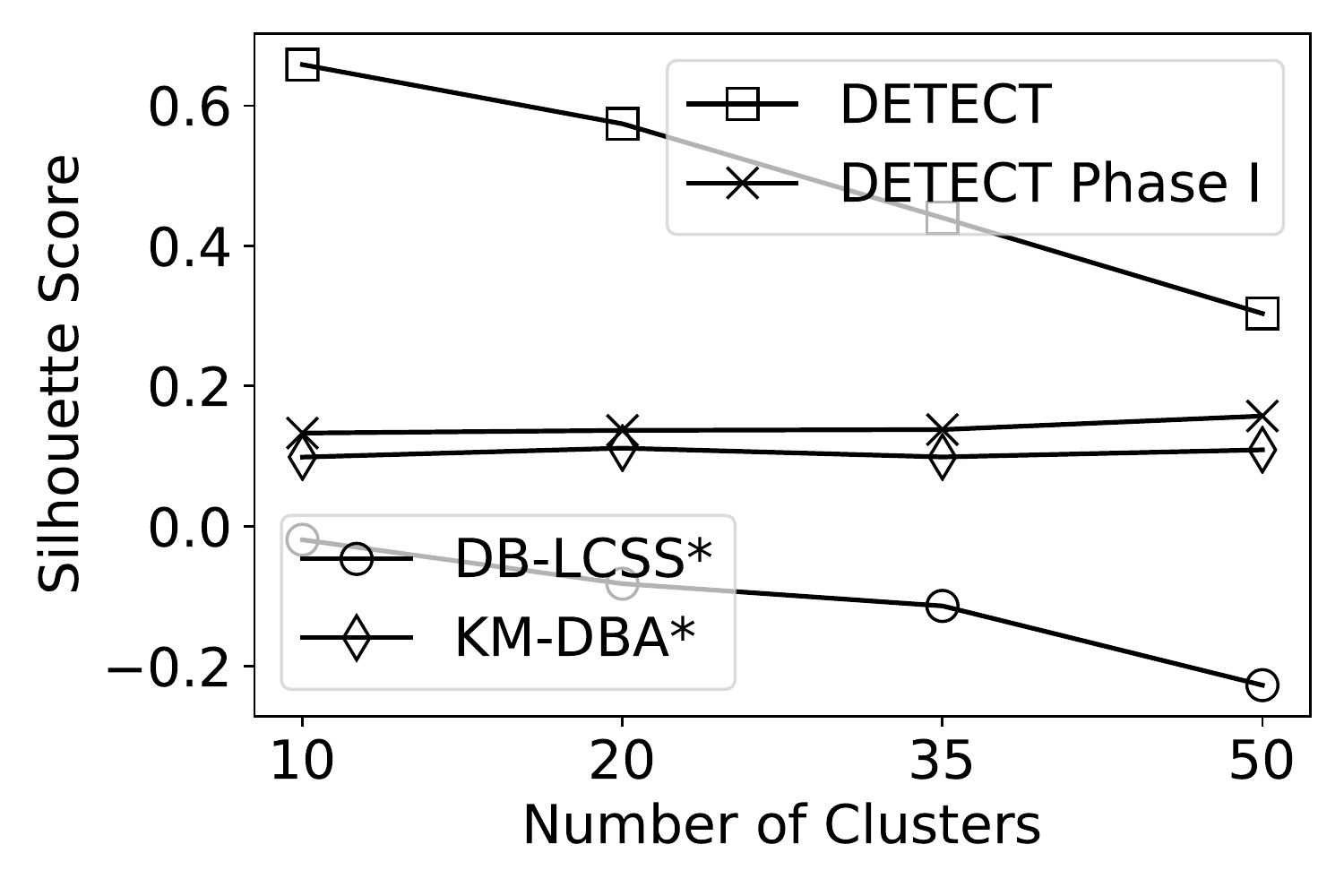}
        \caption{Silhouette Score}
    \end{subfigure}
    \begin{subfigure}[b]{0.47\columnwidth}
    \centering
        \includegraphics[width=\columnwidth]{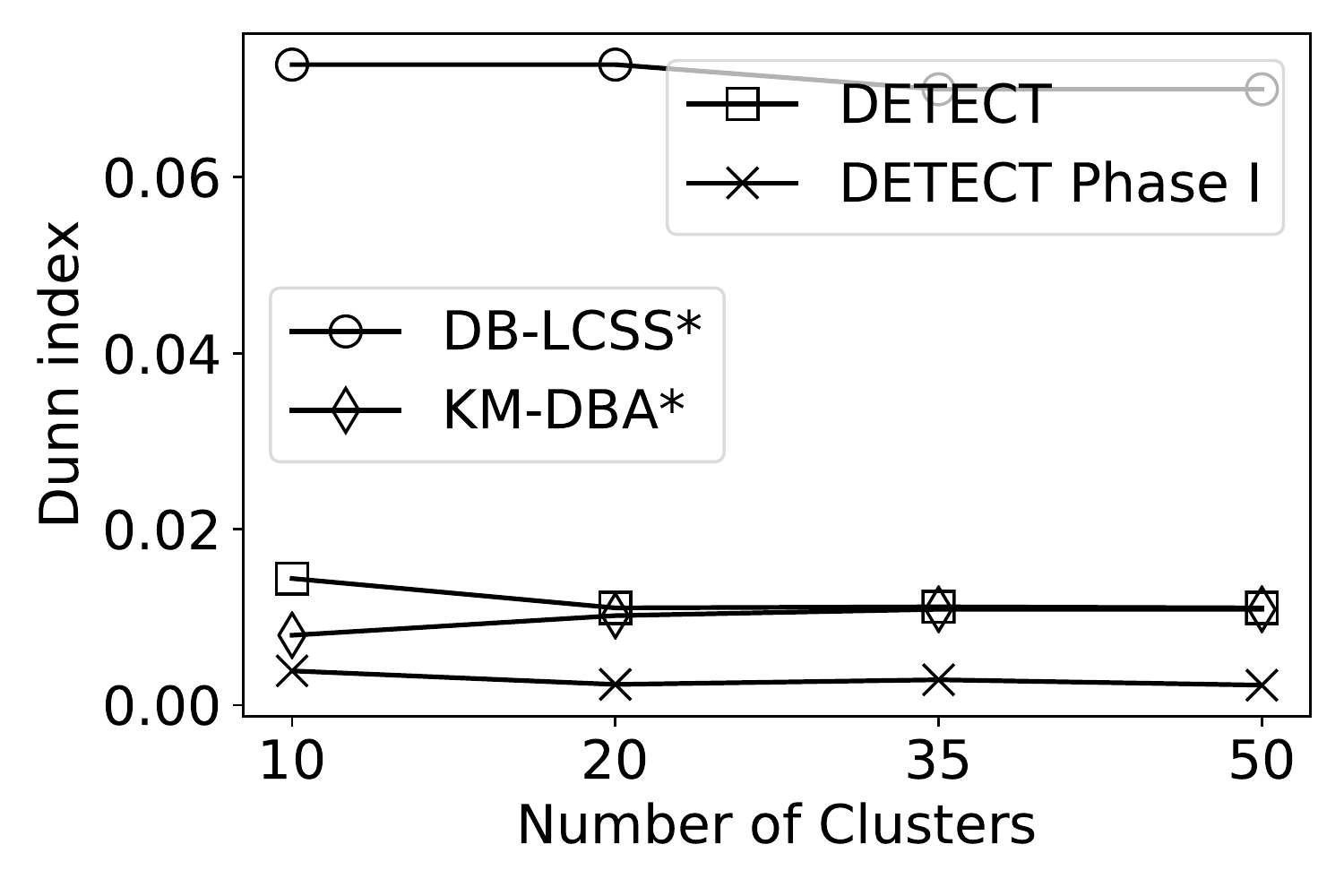}
        \caption{Dunn Index}
    \end{subfigure}
    \begin{subfigure}[b]{0.47\columnwidth}
    \centering
    \includegraphics[width=\columnwidth]{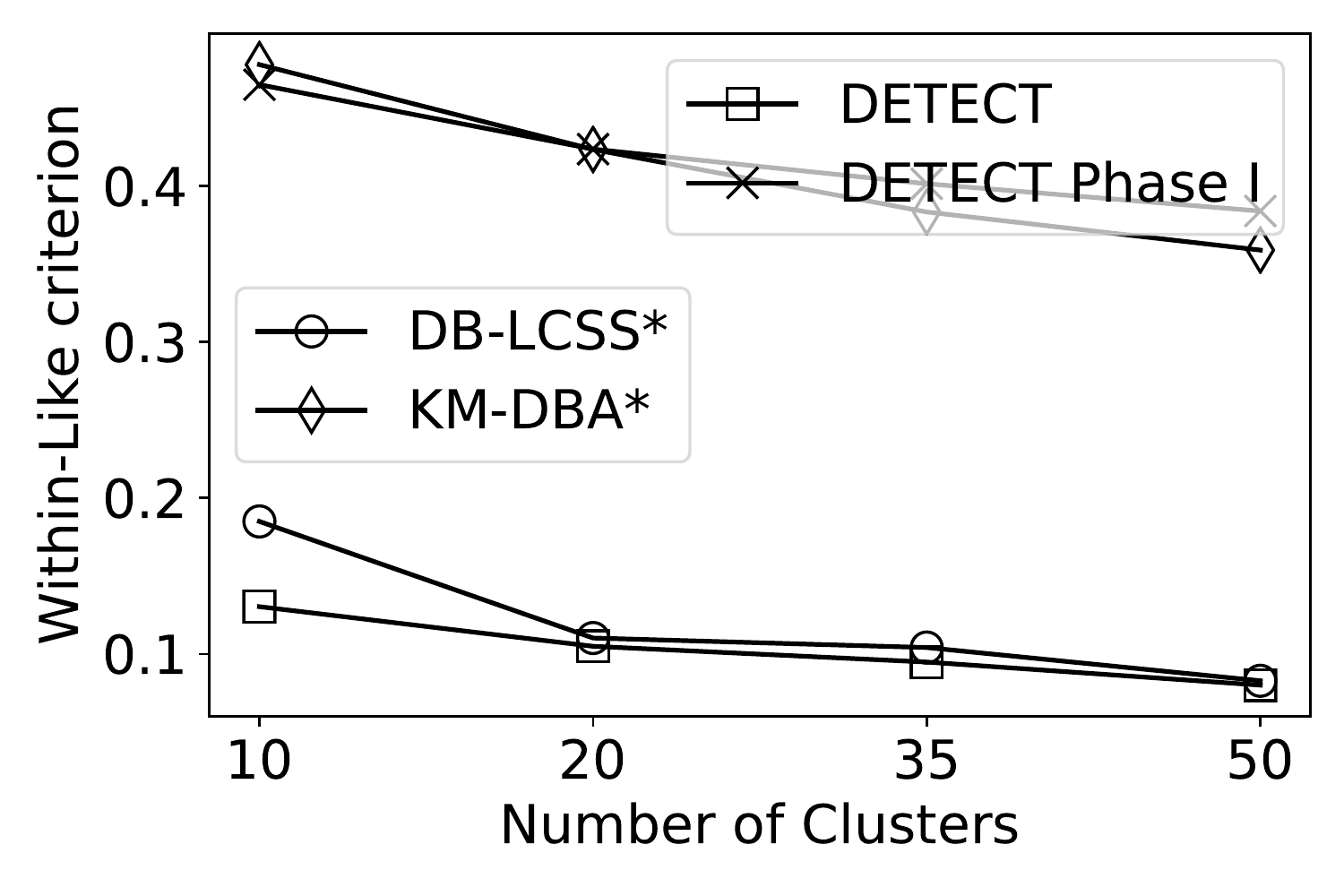}
        \caption{Within-like criterion}
    \end{subfigure}
    \begin{subfigure}[b]{0.47\columnwidth}
    \centering
    \includegraphics[width=\columnwidth]{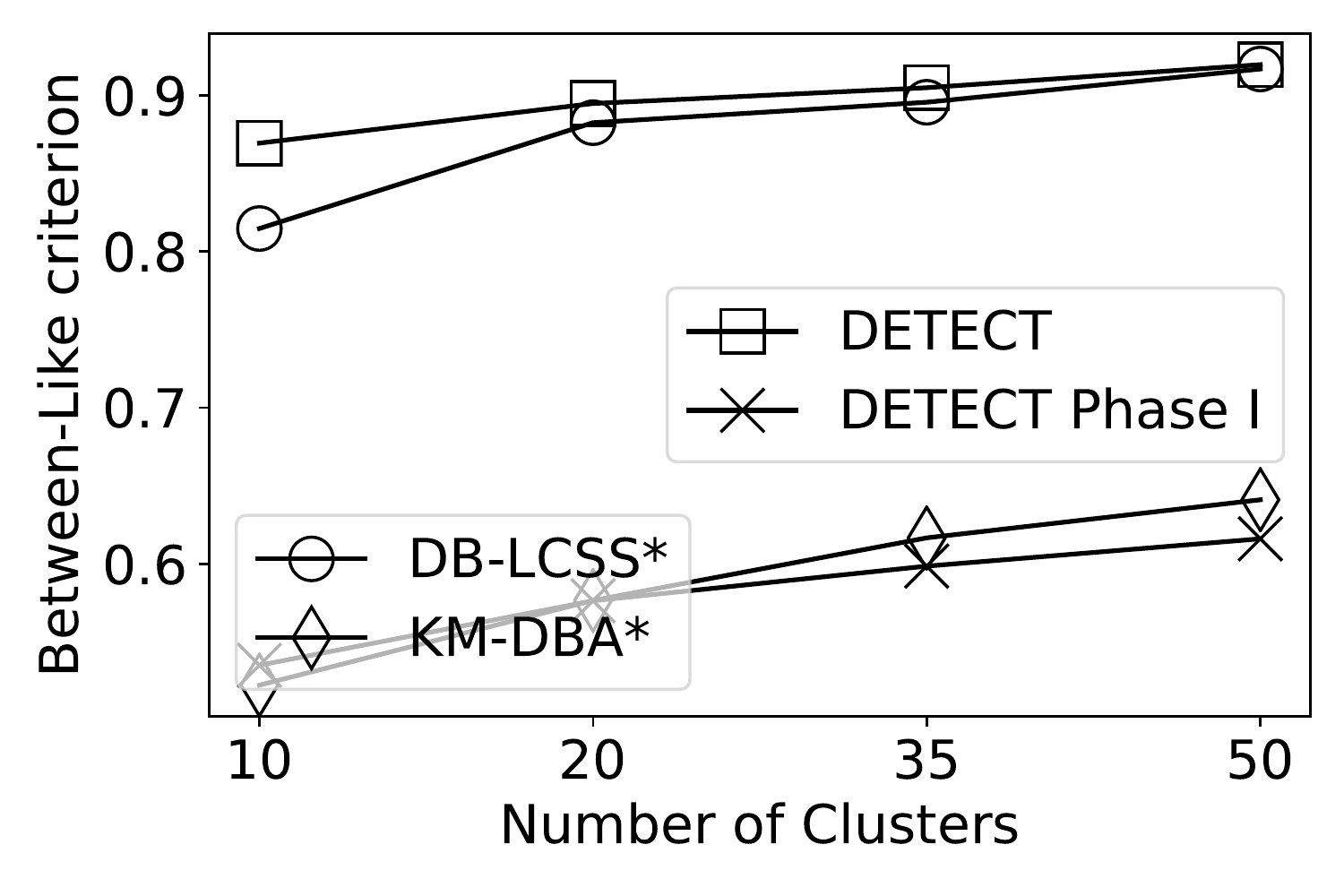}
        \caption{Between-like criterion}
    \end{subfigure}
    \caption{Internal validation of clustering quality.}
    \label{fig:internal}
    \vspace{-10pt}
\end{figure}

\begin{figure}[tbp]
 \centering
\includegraphics[width=0.6\columnwidth]{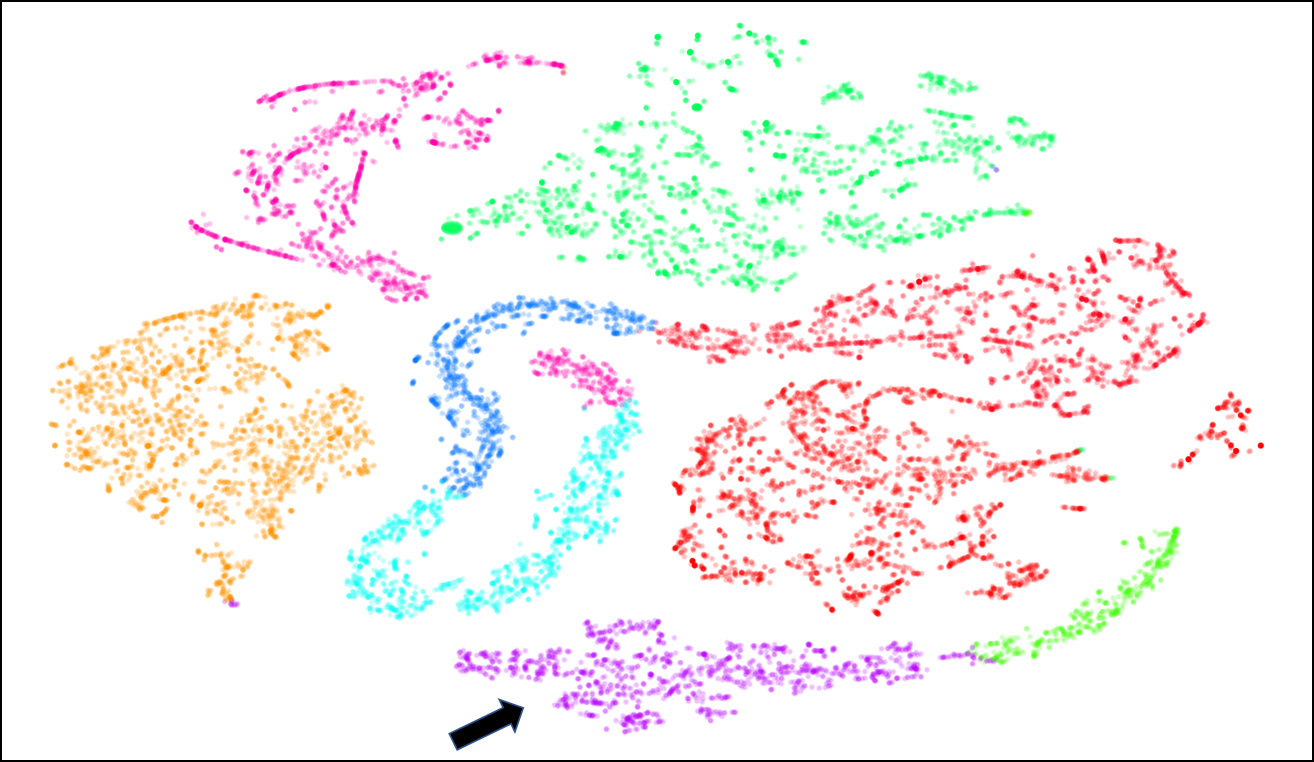}
\caption{Clusters of all trajectories in GeoLife.}
\label{fig:cluster_all}
\vspace{-10pt}
\end{figure}

\begin{figure}[tbp]
 \centering
\includegraphics[width=0.7\columnwidth]{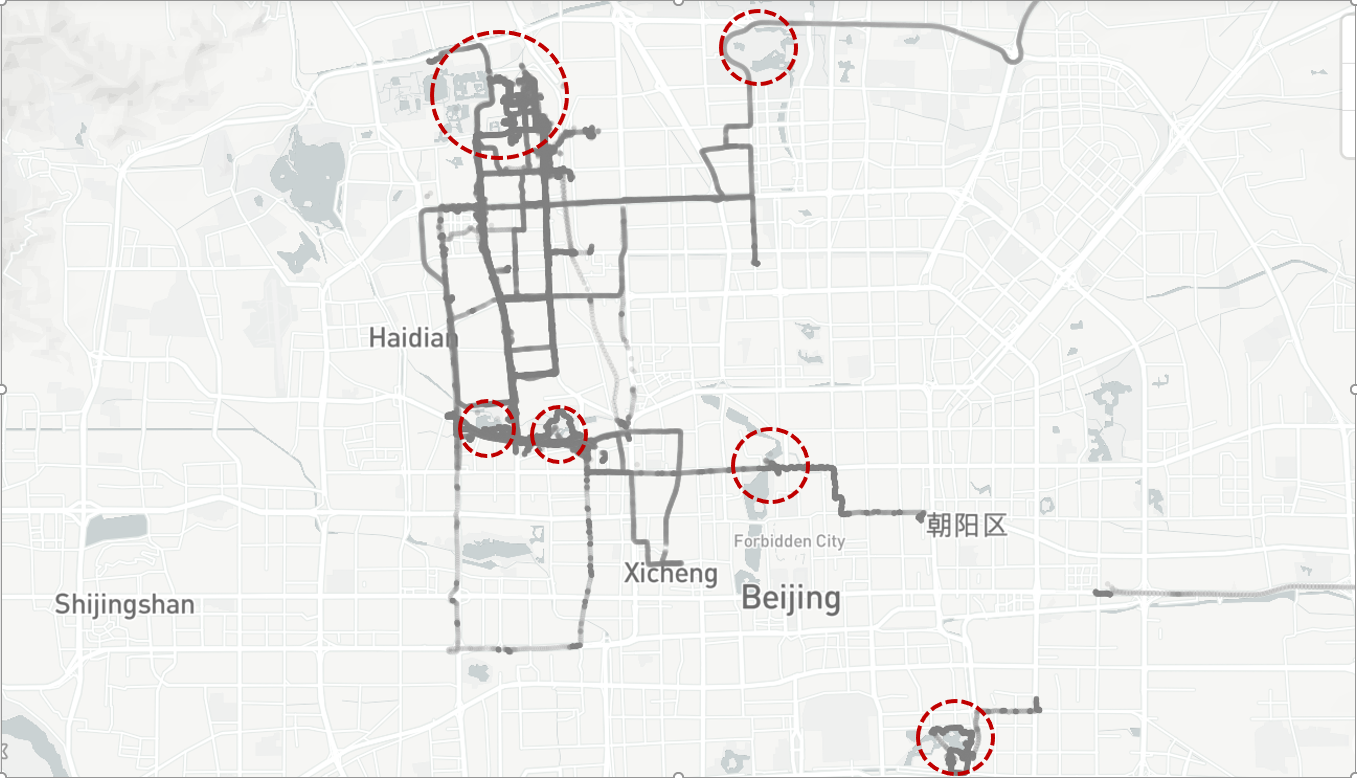}
\caption{Visualization of a detected GeoLife trajectory cluster with locations of recreational parks in red circles.}
\label{fig:cluster_xx}
\vspace{-7pt}
\end{figure}

Figure~\ref{fig:internal} presents the results. All four internal measures indicate that DETECT Phase \RNum{2} increases compactness within (smaller Within-Like) and separation between (larger Between-Like) the clusters generated via Phase \RNum{1}. Moreover, the evolution of Silhouette score, Within-Like criterion and Between-Like criteria suggest that just a few clusters (i.e. 10) can best capture common mobility behaviors. A visual inspection of the 10 clusters (in Figure~\ref{fig:cluster_all}) generated using DETECT establishes its effectiveness in generating well-behaved clusters. Moreover, a closer look at the trajectories in the purple cluster (Figure~\ref{fig:cluster_xx}) illustrates that the discovered mobility behavior can be easily understood. All comprising trajectories---albeit originate from various locations (such as schools and residences)---always involve activities in parks (red dashed circles). 

Lastly, we remark upon the running time of the compared approaches over the full GeoLife dataset. Within our experimental testbed Phase \RNum{1 } trained for 0.8 hour, which was further optimized in Phase \RNum{2} for an additional 1.3 hours. The total computational time of 2.1 hours is still substantially better than the 4.3 hours for KB-DBA*, and 5.2 hours for DB-LCSS*, both of which require $O(n^2)$ pairwise distance computations. 

\section{Related Work} \label{sec:relat}
Trajectory clustering is an important problem and has received significant attention in the past decade. Most techniques utilize raw trajectories (e.g., \cite{yoon2008robust}) with a variety of distance/similarity measurements such as the classic Euclidean, Hausdorff, LCSS, DTW, Frechet, SSPD~\cite{besse2016review} or ad-hoc measurements suited to specific applications (e.g., \cite{lin2008one, ferreira2013vector}). To deal with trajectories of large cardinality, some techniques resort to finding local patterns in sub-trajectories. Lee et al. ~\cite{lee2007trajectory} partition trajectories into minimum description length (MDL) and then cluster the partitioned trajectories. However, these methods operate over raw data points and remain sensitive to the wide range of spatio-temporal scales in human mobility patterns~\cite{yuan2017review}.
Since raw GPS coordinates and timestamps, in and of themselves do not provide semantic information regarding a trajectory, several techniques use movement characteristics derived from the raw data for clustering tasks. In recent work, Yao et. al.~\cite{yao2017trajectory} extract the speed, acceleration, and change of ``rate of turn'' (ROT) of each point in a trajectory as the input sequence for clustering. Wang et al. ~\cite{wang2013semantic} annotate trajectories with movement labels such as ``Enter'', ``Leave'', ``Stop'', and ``Move'' and use these labels to detect events from trajectories. However, just like the approaches based on raw trajectory data, these methods completely fail when dealing with scale-variant trajectories and require carefully extracting movement characteristics, which often do not represent mobility behaviors. Recent work in the privacy literature~\cite{bindschaedler2016synthesizing, gramaglia2017preserving} make efforts to capture mobility semantics so as to anonymize real trajectories or synthesize fake trajectories that are indiscernible from real data. Most related to our work is~\cite{choi2017efficient}, which presents methods to mine patterns in trajectories (eg. work-to-pub). These patterns are limited to places visited one-after-the-other by a user. Hence, they rely on using the check-in point of interest as their data input along with a description of the POI (as opposed to GPS data inputs in our method). In addition, they utilize distance-based clustering to discover \textit{regions}(as opposed to mobility \textit{behaviors}) where these patterns occur frequently.

In neural network literature, solving unsupervised tasks using latent-features learned over deep networks has seen a recent surge. Autoencoders have been used for several unsupervised learning tasks~\cite{xie2016unsupervised,guo2017improved,guo2017deep}. Their inputs are mainly images and texts, which make the techniques inapplicable to the problem of clustering trajectories over mobility behaviors. Sequence-to-sequence autoencoders proposed in \cite{dai2015semi} are applicable to sequential trajectories data, e.g. as used in~\cite{yao2018learning, yao2017trajectory}. However, their main focus is on learning the spatio-temporal properties of trajectories rather than mining the context of the user activities. Lastly, AutoWarp~\cite{abid2018learning} overcomes the problem of manually specifying a specific metric (e.g. LCSS, DTW) for time-series data. It learns a warping distance that mimics latent representations learned via a sequence-to-sequence autoencoder. AutoWarp is especially robust to extremely noisy data.

\section{Conclusion} \label{sec:con} 
We proposed DETECT, a powerful neural framework armed with a clustering oriented loss. We reduced the scale-variance in trajectories using an efficient stay point extracting procedure. These critical parts of the trajectory when augmented with context to capture their geographical influence improve even the baseline KM-DBA by up to 58\%. Furthermore, equipped with input features conducive to mobility behavior analysis, feature augmentation, enables the auto-encoder to embed trajectories in the latent space of behaviors. A joint optimization of the latent embedding improves clustering compactness even further (up to 20\%, measured as the ratio of performance between DETECT and DETECT Phase~\RNum{1}). An exhaustive experimental evaluation confirms the effectiveness of DETECT, consistently achieving at least 40\% improvement over the state-of-the-art baselines in all evaluated external metrics. At the same time, internal validation measures and running time efficiency results mean that DETECT is also scalable to large datasets. For future work, a promising direction is to explore how learned clusters can be used as features in supervised learning models.
\section{Acknowledgement}
This material is based on research supported in part by the USC
Integrated Media Systems Center.
Any opinions, findings, and conclusions or recommendations expressed in this material are those of the authors and do
not necessarily reflect the views of any of the sponsors.

\bibliographystyle{abbrv}
\bibliography{citation}

\end{document}